\PassOptionsToPackage{normalem}{ulem}
\documentclass[]{arxiv_template}

\usepackage{enumitem}
\usepackage[toc,page,header]{appendix}
\usepackage[utf8]{inputenc}
\usepackage[T1]{fontenc}
\usepackage{url}
\usepackage{float}
\usepackage{makecell}
\usepackage{microtype}
\usepackage{graphicx}
\usepackage{xspace}
\usepackage{colortbl}
\usepackage{booktabs}
\usepackage{multirow}
\usepackage{amsmath,amssymb}
\usepackage{algorithm}
\usepackage{algpseudocode}
\usepackage{tabularx}
\usepackage{array}
\usepackage{threeparttable}
\usepackage[export]{adjustbox}
\usepackage[most]{tcolorbox}
\usepackage{listings}
\usepackage{needspace}
\usepackage{extramarks}
\tcbuselibrary{listings,breakable,skins}
\usepackage{lineno}
\definecolor{oursbg}{RGB}{239,246,255}
\definecolor{headerbg}{RGB}{246,247,249}
\definecolor{good}{RGB}{32,120,74}
\definecolor{bad}{RGB}{176,55,55}
\definecolor{rulegray}{RGB}{170,174,181}


\lstdefinestyle{promptstyle}{
  basicstyle=\fontsize{6.7pt}{7.2pt}\selectfont\ttfamily,
  breaklines=true,
  breakatwhitespace=false,
  breakindent=0pt,
  columns=fullflexible,
  keepspaces=true,
  showstringspaces=false,
  upquote=true,
  extendedchars=true,
  postbreak=\mbox{{\color{gray}$\hookrightarrow$}\space},
  aboveskip=0pt,
  belowskip=0pt,
  xleftmargin=0pt,
  xrightmargin=0pt,
  literate=
    {—}{{---}}1
    {–}{{--}}1
    {“}{{``}}1
    {”}{{''}}1
    {‘}{{`}}1
    {’}{{'}}1,
}

\lstdefinestyle{codestyle}{
  language=Python,
  basicstyle=\footnotesize\ttfamily,
  keywordstyle=\color{blue!70!black}\bfseries,
  commentstyle=\color{green!40!black}\itshape,
  stringstyle=\color{purple!70!black},
  numberstyle=\tiny\color{gray},
  backgroundcolor=\color{gray!8},
  frame=single,
  rulecolor=\color{black!40},
  framesep=4pt,
  framerule=0.4pt,
  breaklines=true,
  breakatwhitespace=false,
  breakindent=0pt,
  columns=fullflexible,
  keepspaces=true,
  showstringspaces=false,
  upquote=true,
  extendedchars=true,
  postbreak=\mbox{{\color{gray}$\hookrightarrow$}\space},
  aboveskip=6pt,
  belowskip=6pt,
  xleftmargin=4pt,
  xrightmargin=4pt,
  literate=
    {—}{{---}}1
    {–}{{--}}1
    {“}{{``}}1
    {”}{{''}}1
    {‘}{{`}}1
    {’}{{'}}1,
}

\newtcblisting{promptbox}[2][]{
  enhanced jigsaw,
  breakable,
  colback=gray!4,
  colframe=black!55,
  boxrule=0.4pt,
  arc=1.5pt,
  left=2pt,
  right=2pt,
  top=2pt,
  bottom=2pt,
  fonttitle=\bfseries\small,
  coltitle=white,
  colbacktitle=black!60,
  attach boxed title to top left={yshift=-2pt,xshift=4pt},
  boxed title style={colback=black!60,boxrule=0pt,arc=1pt},
  title=#2,
  listing only,
  listing options={style=promptstyle},
  #1,
}

\newcommand{\approach}{{SWE-Pruner Pro}\xspace}

\newcommand{\improve}[1]{\,\textcolor{good}{\scriptsize$\uparrow$#1}}
\newcommand{\reduce}[1]{\,\textcolor{good}{\scriptsize$\downarrow$#1}}
\newcommand{\worsen}[1]{\,\textcolor{bad}{\scriptsize$\uparrow$#1}}
\newcommand{\loss}[1]{\,\textcolor{bad}{\scriptsize$\downarrow$#1}}

\renewcommand{\title}[1]{%
  \gdef\thetitle{#1}%
  \gdef\titlelist{\Large\rmfamily\bfseries #1}%
}

\newcommand{\shortpapertitle}{SWE-Pruner Pro}
\fancypagestyle{titlestyle}{
  \fancyhead[L]{\small\sffamily\color{epiccolor}\firstleftmark}
  \fancyhead[R]{\small\sffamily\color{epiccolor}\shortpapertitle}
  
  \renewcommand{\headrule}{%
    \vskip -2mm
    \hbox to\headwidth{\color{epiccolor}\leaders\hrule height 0.5pt\hfill}}
}

\title{SWE-Pruner Pro: The Coder LLM Already Knows What to Prune}

\makeatletter
\gdef\authorlist{%
  \begin{center}
    \normalfont\rmfamily\bfseries\fontsize{9.8pt}{11.8pt}\selectfont
    Yuhang Wang$^{1,*}$,
    Yuling Shi$^{1,*}$,
    Shaoqiu Zhang$^{1}$,
    Jialiang Liang$^{1}$,
    Shilin He$^{2}$\\
    Siyu Ye$^{2}$,
    Yuting Chen$^{1}$,
    Kai Cai$^{2}$,
    Xiaodong Gu$^{1,\dagger}$
  \end{center}
}
\gdef\affiliationlist{%
  \begin{center}
    \normalfont\rmfamily\fontsize{8.8pt}{10.6pt}\selectfont
    $^{1}$LLM4SE Lab, Shanghai Jiao Tong University\\
    $^{2}$Douyin Group
  \end{center}
}
\makeatother

\abstract{
Pruning long context for coding agents has been a vital technology for efficient context management. While existing context pruning methods such as SWE-Pruner realize this by attaching a separate code classifier, we find the agent itself encodes internal representations indicating the relevance of code context when reading tool output. Based on this finding, we propose \approach, which prunes tool outputs directly inside the agent. Concretely, a small head turns the agent's own internal representations into a keep-or-prune label for each line, with a length-aware embedding keyed to each tool output's line count. Across two open-weight backbones and four multi-turn benchmarks, \approach saves up to 39\% of prompt and completion tokens while preserving task quality, with bounded inference overhead. Notably, on \textsc{MiMo-V2-Flash} \approach additionally raises the SWE-Bench Verified resolve rate by $+3.8\%$ and the long-context Oolong accuracy by $+2.2$ points.
}

\titleformat*{\paragraph}{\bfseries}
\checkdata[Project Page]{\url{https://github.com/Ayanami1314/swe-pruner-pro}}

\begin{document}
\maketitle
\pagestyle{titlestyle}
\tcbset{reset}

{
\renewcommand{\thefootnote}{\fnsymbol{footnote}}
  \footnotetext[1]{Equal contribution.}
  \footnotetext[2]{Corresponding author: \texttt{xiaodong.gu@sjtu.edu.cn}.}
}

\section{Introduction}
\label{sec:intro}
Pruning long context for coding agents has been a vital technology for efficient context management \cite{cheng2024xrag,shi2025longcodezip,wang2026survey}. During programming tasks, coding agents accumulate very long tool outputs full of redundant information during multi-turn interaction with environments. They solve repository-level tasks~\citep{DBLP:journals/corr/abs-2608-09802} by interleaving reasoning with environment interactions: they call tools like \texttt{cat}, \texttt{grep}, \texttt{ls}, and \texttt{python}, observe the raw textual output, and continue. In practice, the bulk of the per-trajectory token budget is spent on tool outputs, much of it repetitive or never re-referenced~\citep{swepruner2026}. This redundant content accumulates across turns, inflating cost and triggering well-documented long-context degradation~\citep{liu2023lost,laban2025llms,li2023loogle}.
As such, pruning the long context has been a natural way to keep the context manageable. 

Two lines of prior work address this. The first, general-purpose prompt or code compression~\citep{jiang2023llmlingua,li2023compressing,shi2025longcodezip}, scores tokens with a fixed metric such as perplexity or syntactic structure, and so cannot adapt to the agent's evolving focus during the task. The second, task-specific pruning, exemplified by \mbox{SWE-Pruner}~\citep{swepruner2026}, does condition on the agent's intent, but at the cost of a second scoring model and an explicit goal-hint query the agent must write every turn.

\begin{figure}[!t]
\centering
\includegraphics[
  max width=0.90\linewidth,
  max totalheight=0.27\textheight
]{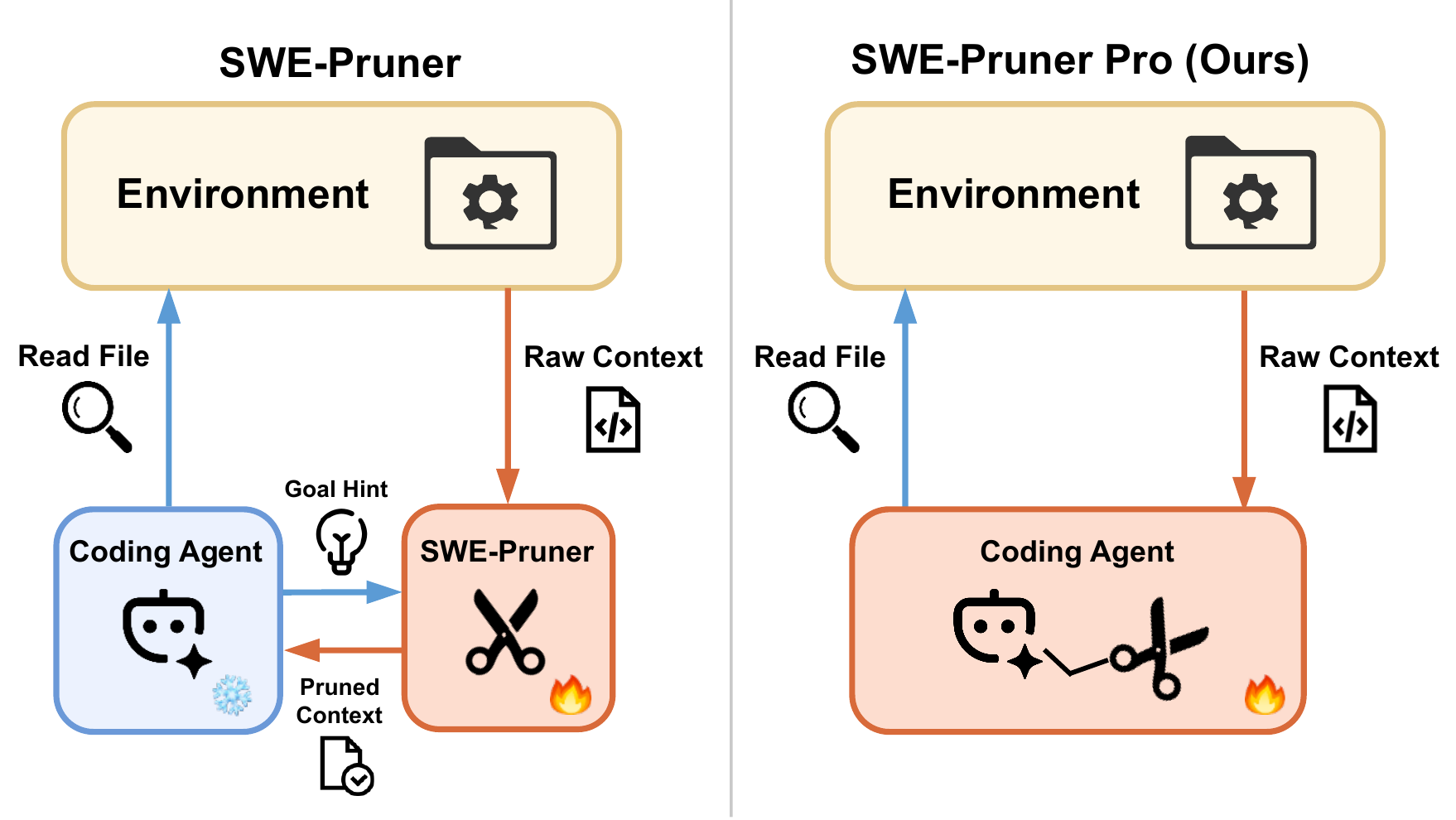}
\caption{(Left) Prior work~\citep{swepruner2026} uses a separate \mbox{Pruner} model conditioned on an explicit goal-hint query the agent must write at every turn. (Right) \approach prunes directly from the agent's own internal representations, with no external scoring model.}
\label{fig:teaser}
\end{figure}


Both approaches obtain the pruning signal from outside the agent, even though the agent's backbone model has already processed the tool output. We ask whether the information needed to decide which lines to keep or prune is already present in the model's own representations, motivating us to read the pruning signal directly from them rather than eliciting an additional goal-hint query~\citep{swepruner2026}, as illustrated in Figure~\ref{fig:teaser}. Our probing study shows that kept and pruned lines are distinguishable in this representation space, with even a simple linear probe providing evidence that per-line pruning information is already there (\S\ref{sec:motivation}).


We instantiate this idea as \approach, which turns the agent's own representations into line-level pruning decisions. When the agent reads a tool response, its frozen backbone already produces token representations during the normal prefill; \approach reuses these representations to predict which lines should be kept or pruned. A lightweight head makes these predictions, with two additions that matter in practice: length awareness, so the head can behave differently on short and long outputs, and a per-sample balanced focal loss, which rebalances each sample's keep and prune tokens equally, so the rarer the kept lines are, the more importance signal they carry into the objective. Because the head runs in-server on the existing prefill, it avoids an extra model call and adds only a small bounded cost.

Across two open-weight backbones and four multi-turn benchmarks, \approach is the most consistent pruner among the seven methods we evaluate. On the SWE-QA family and Oolong, it is the only method that reduces end-to-end token use in every setting while keeping task quality close to the unpruned agent. The savings reach 39\% on SWE-QA-Pro and 30\% on the long-context Oolong benchmark, where \approach also improves \textsc{MiMo-V2-Flash} accuracy by $+2.2$ points. On the coding-agent benchmark SWE-Bench Verified, \approach improves \textsc{MiMo-V2-Flash}'s resolve rate by $+3.8\%$. In our replay study, the in-server head adds only 8.4\% aggregate wall time on top of the agent's total generation time, without requiring an extra model call.

Our contributions are:
\begin{itemize}
    \item We show that an agent backbone's internal representations already encode line-level importance of tool outputs, removing the need for a separate scoring model or explicit query.
    \item We propose \approach, a lightweight head that reads the pruning signal directly from the backbone with a learned length-aware embedding and a per-sample balanced focal loss.
    \item We evaluate \approach on two open-weight backbones and four multi-turn benchmarks, where it saves up to 39\% of tokens while preserving task quality.
\end{itemize}

\section{Motivation}
\label{sec:motivation}

\begin{figure}[!t]
  \centering
  \begin{minipage}[t]{0.49\linewidth}
    \centering
    \includegraphics[width=\linewidth,height=0.225\textheight,keepaspectratio]{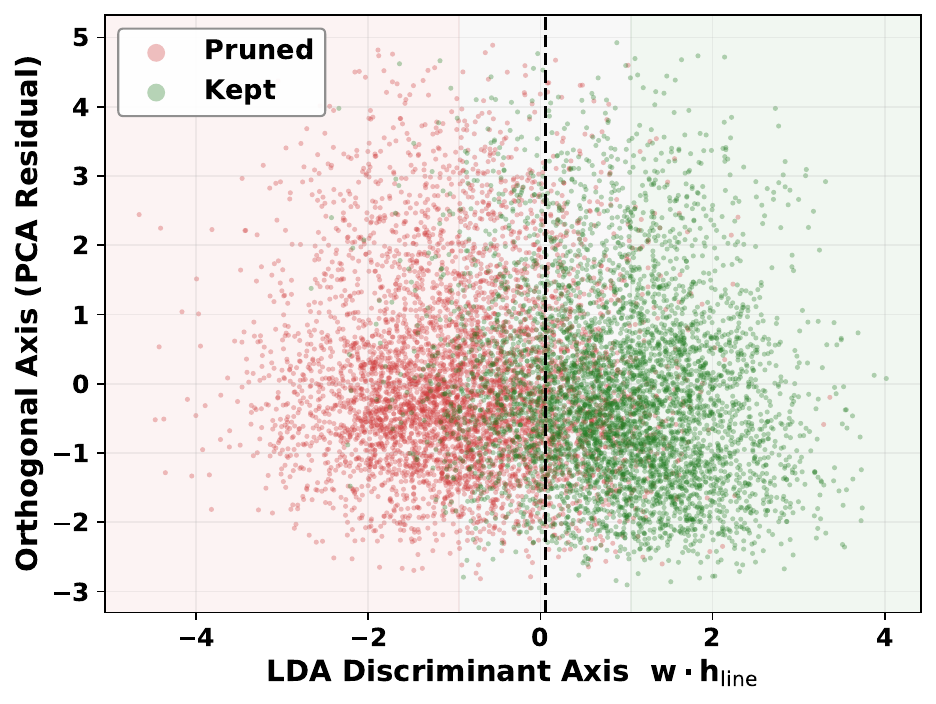}
  \end{minipage}\hfill
  \begin{minipage}[t]{0.49\linewidth}
    \centering
    \includegraphics[width=\linewidth,height=0.225\textheight,keepaspectratio]{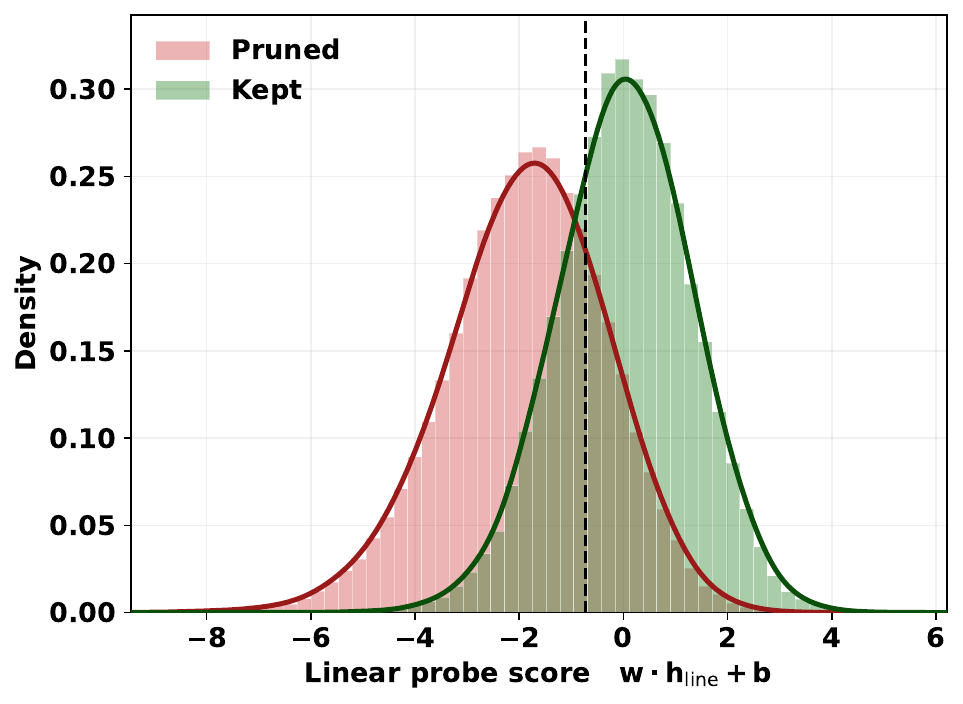}
  \end{minipage}
  \caption{Linear probe on the agent backbone's frozen last-layer hidden states. (Left) Two-dimensional projection along the LDA discriminant axis and an orthogonal direction. (Right) Distribution of probe scores. The two classes have visibly different means but overlap in the middle band.}
  \label{fig:linear-probe}
\end{figure}

Multi-turn coding agents spend the bulk of their per-trajectory token budget on tool outputs. On SWE-Bench Verified, file-reading commands account for over 70\% of the tokens consumed by Mini-SWE-Agent backed by Claude Sonnet 4.5, with a comparable pattern observed on GLM-4.6~\citep{swepruner2026}; this cost compounds across turns as earlier reads remain in the context window. Existing pruners reduce this cost by adding machinery on top of the agent. General-purpose compressors~\citep{jiang2023llmlingua,li2023compressing,shi2025longcodezip} score tokens by a fixed surrogate such as perplexity, with no view of the agent's task; task-specific pruners~\citep{swepruner2026} run a separate scoring model conditioned on an explicit goal-hint query the agent writes at every turn. Both routes treat the agent's current information need as a quantity to be reconstructed externally, rather than read directly from the backbone that is already processing the tool output.

Reading a tool output is itself an attention-weighted forward pass over its tokens, so the backbone must already encode which lines matter for the next action and which it can ignore. If so, the keep-or-prune signal should already be visible in its last-layer hidden states. To check this, we collect $\approx 2{,}260$ multi-turn tool responses ($\approx 155$k lines) from publicly released SWE-Bench-style and terminal-task datasets, and label each line as keep or prune with \textsc{Claude Sonnet~4.6}; trajectory sources and the full labelling protocol are in Appendix~\ref{app:training-data}. We then freeze \textsc{Qwen3-Coder-Next}, mean-pool the last-layer hidden states of each line, and fit a logistic regression to see whether the two classes are separable.

The two classes have visibly different means in the hidden-state space: along the LDA discriminant axis their distributions are offset (Figure~\ref{fig:linear-probe} left), and the per-line scores along the same direction confirm a clear mean gap with substantial overlap in the middle band (Figure~\ref{fig:linear-probe} right). On the held-out trajectories, the probe reaches AUC $0.83$ and best-$F_1$ $0.63$, well above the majority-class $F_1$ upper bound of $0.46$ at the empirical positive rate of ${\approx}30\%$. The keep-or-prune signal is therefore already inside the backbone's representation. A logistic regression alone cannot resolve the overlap in the mid-score region of Figure~\ref{fig:linear-probe} (right) or exploit the length-dependent structure of tool outputs; both motivate \approach's non-linear head.

\FloatBarrier
\section{Method}
\label{sec:method}

\begin{figure}[!t]
\centering
\includegraphics[
  max width=0.90\linewidth,
  max totalheight=0.43\textheight
]{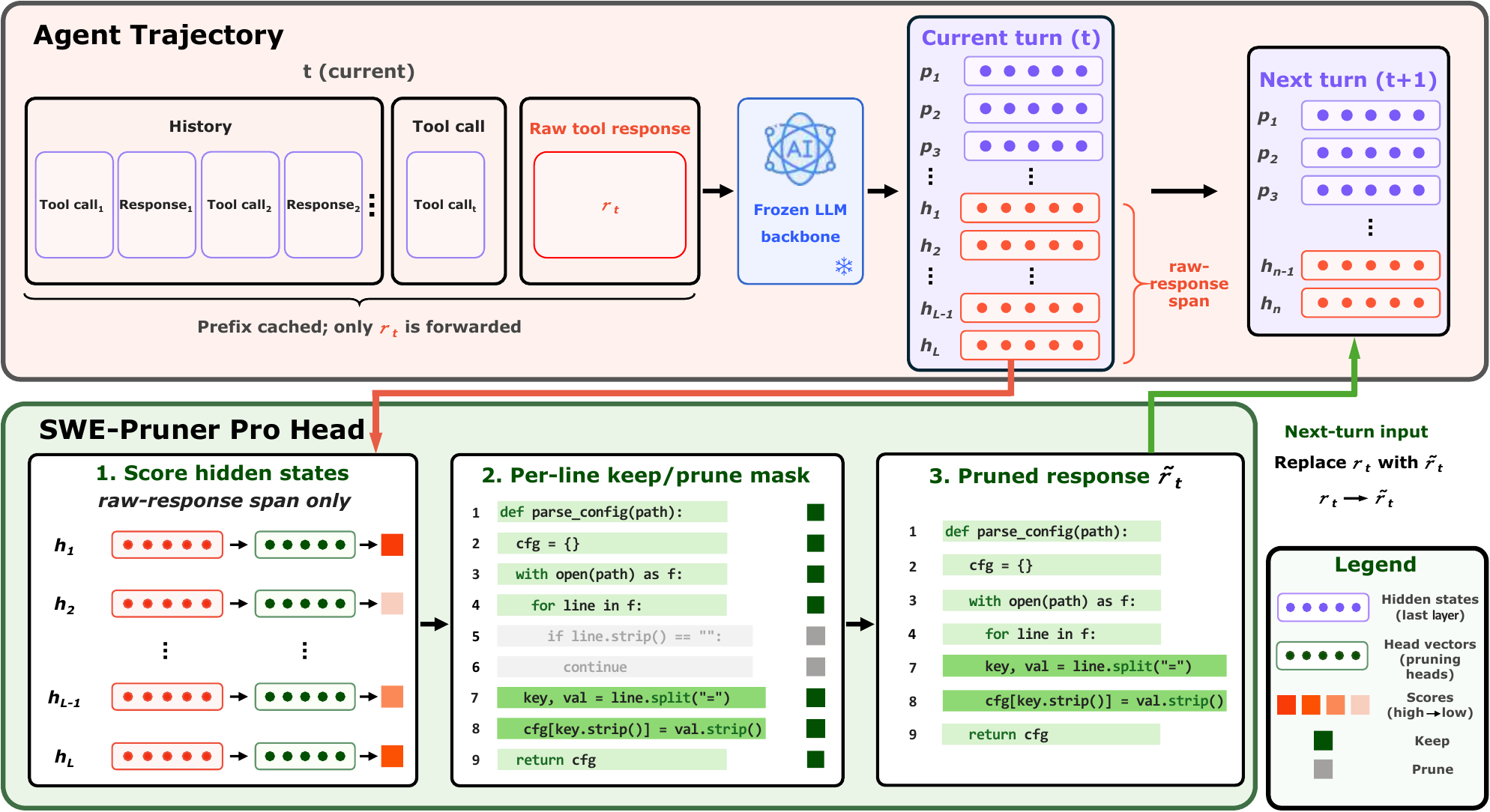}
\caption{Overview of \approach. \textbf{Top (Agent Trajectory):} at each turn the agent backbone prefills $[\text{history}, c_t, r_t]$; only the new tool-response tokens are forwarded, and \approach reads their last-layer hidden states $\{h_i\}$ off this prefill at no extra forward on $r_t$. \textbf{Bottom (\approach head):} the head scores each $h_i$, aggregates the binarised scores into per-line keep-or-prune decisions, and the pruned response $\tilde r_t$ replaces $r_t$ in the next turn.}
\label{fig:overview}
\end{figure}

\approach makes per-line keep-or-prune decisions on each tool response directly from the agent backbone's last-layer hidden states (Figure~\ref{fig:overview}). The remainder of this section covers the per-turn pipeline (\S\ref{subsec:pipeline}), the pruning head (\S\ref{subsec:head}), and training (\S\ref{subsec:training}).

\subsection{Pipeline overview}
\label{subsec:pipeline}
At each turn $t$, the agent issues a tool call $c_t$ and the environment returns a raw tool response $r_t$, often hundreds of lines long and persisting in the agent's context for the rest of the trajectory. Before the agent generates its next move, the backbone prefills $[H_{t-1}, c_t, r_t]$ into its KV cache (Figure~\ref{fig:overview} top), where $H_{t-1}$ is the context history accumulated before turn $t$. The prefix $[H_{t-1}, c_t]$ is already in the KV cache, so only the new $r_t$ tokens are forwarded, producing last-layer hidden states $h_1, \ldots, h_L$ over the $r_t$ span, where $L = |r_t|$ in tokens.

\approach attaches at this prefill. The pruning head turns each $h_i$ into a per-token keep-or-prune logit and aggregates them into per-line decisions (Figure~\ref{fig:overview} bottom). Line-level granularity preserves the syntactic structure of the kept code while still allowing fine-grained pruning, consistent with prior code-aware pruners~\citep{shi2025longcodezip,swepruner2026}. Pruning is applied between turns: the agent's own generation at turn $t$ still attends to the full $r_t$, while $\tilde r_t$ replaces $r_t$ when the trajectory continues into turn $t{+}1$. The head reads $\{h_i\}$ off the prefill the backbone already performs on $r_t$. The only added backbone work is a single re-forward of $\tilde r_t$ at the next turn; since $\tilde r_t$ is typically much shorter than $r_t$ (mean labelled keep-rate ${\approx}30\%$, Appendix~\ref{app:training-data}), this overhead is more than offset by the shorter context all subsequent generation must attend to. The full per-turn loop is given as pseudocode in Algorithm~\ref{alg:perturn}; engineering details for the in-server integration are in Appendix~\ref{app:hs-extraction}.

\subsection{Pruning head}
\label{subsec:head}

\begin{figure}[!t]
\centering
\includegraphics[
  max width=0.86\linewidth,
  max totalheight=0.42\textheight
]{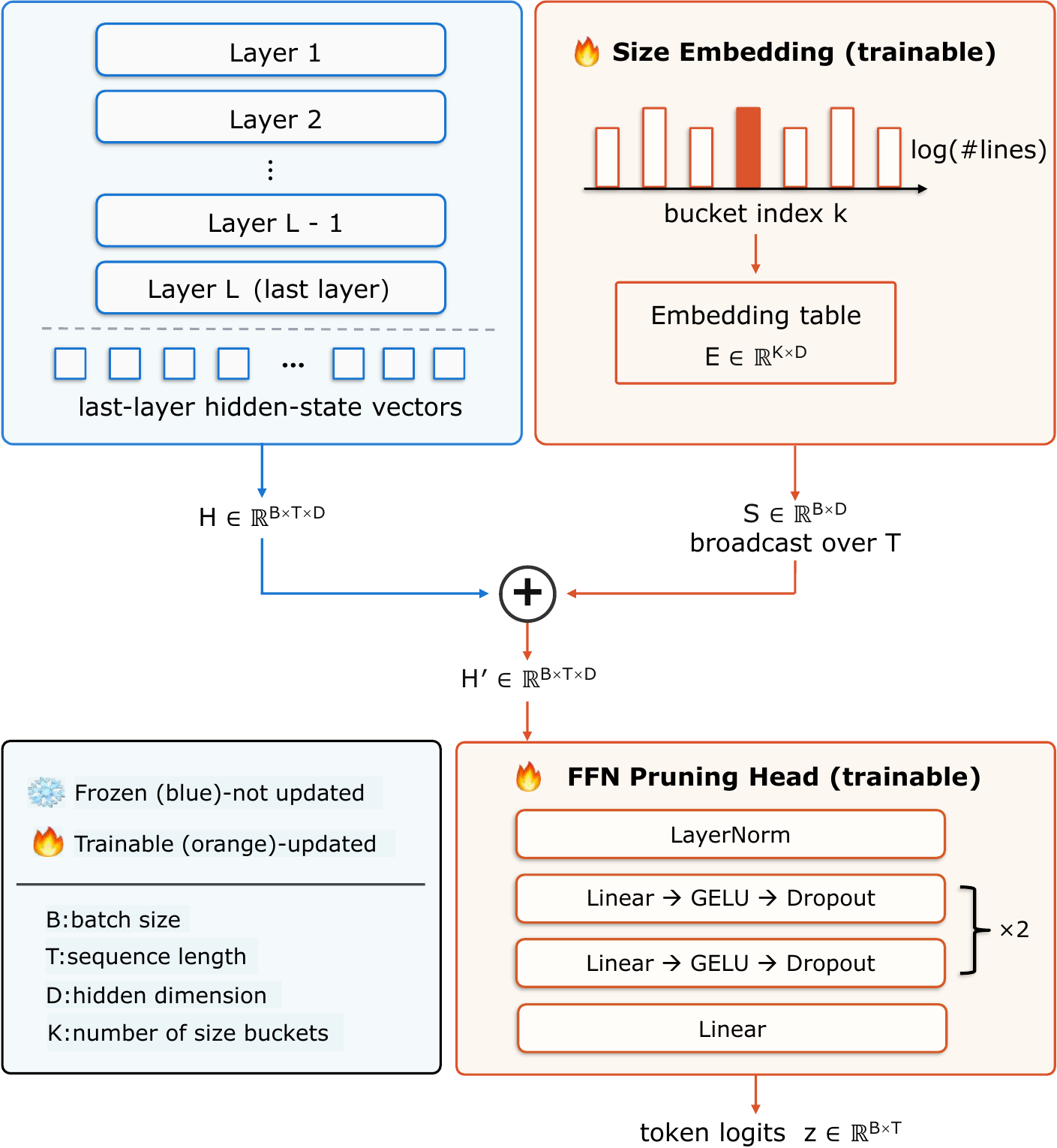}
\caption{Head architecture. The frozen backbone's hidden states $h_i$ receive an additive length-aware embedding indexed by the line count $N$, then pass through a two-block feed-forward stack with LayerNorm and dropout to a per-token keep logit. Line decisions are produced by majority vote within each line at inference.}
\label{fig:head}
\end{figure}

Let $N$ be the number of lines in $r_t$ and let $d$ denote the dimension of $h_i$. The head has two components: a length-aware embedding $\mathbf{e}(N) \in \mathbb{R}^d$ that conditions the keep decision on response length, and a per-token feed-forward classifier $f_\theta$ that maps each augmented hidden state to a keep logit.


\paragraph{Length-aware embedding.}
$\mathbf{e}(N)$ is a learned function of the line count $N$, broadcast-added to every hidden state before $f_\theta$ and zero-initialised so $f_\theta$ reduces to its length-agnostic limit at the start of training:
\begin{equation}
    \tilde{h}_i = h_i + \mathbf{e}(N).
    \label{eq:size-bias}
\end{equation}
The embedding gives the head explicit access to $N$ alongside the per-token hidden state, so its keep-or-prune mapping can vary with response length. This matters because the cost of mis-pruning is highly non-uniform in $N$: a few lines wrongly stripped from a 5-line response is catastrophic, while the same on a 300-line response is negligible. The length-aware embedding lets the head absorb this dependence rather than relying on a single length-agnostic mapping for all lengths.

\paragraph{Per-token classifier.}
$f_\theta$ is a small non-linear network rather than a single linear layer: the linear-probe evidence in \S\ref{sec:motivation} already shows the signal is largely decodable from $h_i$, but a linear classifier alone cannot close the mid-score overlap (Figure~\ref{fig:linear-probe}) or interact with the length-aware embedding of Eq.~\ref{eq:size-bias}; both gaps motivate the small non-linear head. Concretely, $f_\theta$ is a per-token feed-forward classifier: LayerNorm, two hidden Linear-GELU-Dropout blocks with hidden width $d$ matching the backbone's hidden size, and a final Linear projection to a single logit. Applied to every augmented hidden state, it produces keep probabilities:
\begin{equation}
    z_i = f_\theta(\tilde{h}_i), \qquad p_i = \sigma(z_i).
    \label{eq:logit}
\end{equation}

\paragraph{Line-level decision.}
With per-token sigmoid threshold $\tau$ (we use $\tau = 0.5$), per-line decisions $\hat y_\ell$ are obtained by majority vote of the binarized token decisions within each line $\ell$:
\begin{equation}
    \hat{y}_\ell = \mathbb{1}\!\left[\frac{1}{|\ell|}\sum_{i \in \ell} \mathbb{1}[p_i > \tau] > \tfrac{1}{2}\right].
    \label{eq:majority}
\end{equation}
Deferring aggregation to inference time allows the head to be trained from per-token labels expanded directly from per-line annotations. Lines with $\hat{y}_\ell = 0$ are removed from the tool response before it is appended to the agent's history.

\subsection{Training}
\label{subsec:training}
We train the head on $22{,}609$ samples from real multi-turn agent trajectories, each annotated by \textsc{Claude Sonnet~4.6} with per-line keep-or-prune labels.
We expand line labels to per-token labels for the loss, retaining uncertain rows as positives following the labelling protocol in Appendix~\ref{app:training-data}.

The loss is a per-sample balanced focal loss.
Per-line keep/prune labels from an LLM annotator are inherently fuzzy, but the \emph{ratio} the annotator settles on is itself informative: when only 3 of 100 lines are kept, those 3 lines carry the strongest signal about what is irreplaceable; when 90 of 100 are kept, the 10 pruned lines carry the strongest signal about what is safely removable. Cross-entropy and batch-level focal loss see only the global keep rate and treat every token equally, so they overfit the typical keep rate (${\approx}30\%$) and dilute exactly the extreme-ratio samples whose labels carry the most importance. Where focal sharpens the loss around hard decision boundaries, our per-sample balancing instead protects the recall of each sample's minority class: we compute the loss separately over keep and prune tokens within a sample and average the two with equal weight, so each sample contributes equally from each of its classes regardless of its own keep rate.

For each token $i$ in sample $s$, let $p_{t,i} = p_i$ if $y_i = 1$ and $p_{t,i} = 1 - p_i$ otherwise; then:

\begin{equation}
    \mathcal{L}^{\mathrm{tok}}_i = (1 - p_{t,i})^{\gamma} \cdot \mathrm{BCE}(p_i,\, y_i), \quad \gamma = 2.
    \label{eq:focal}
\end{equation}
We average separately over keep and prune tokens within each sample:
\begin{equation}
    \mathcal{L}^{\mathrm{keep}}_s = \frac{\sum_i y_i\, \mathcal{L}^{\mathrm{tok}}_i}{\sum_i y_i}, \quad
    \mathcal{L}^{\mathrm{prune}}_s = \frac{\sum_i (1-y_i)\, \mathcal{L}^{\mathrm{tok}}_i}{\sum_i (1-y_i)},
    \label{eq:balance}
\end{equation}
and combine with equal weights:
\begin{equation}
    \mathcal{L}_s = \tfrac{1}{2}\,\mathcal{L}^{\mathrm{keep}}_s + \tfrac{1}{2}\,\mathcal{L}^{\mathrm{prune}}_s.
    \label{eq:loss-sample}
\end{equation}
The final objective is $\mathcal{L} = \frac{1}{S}\sum_s \mathcal{L}_s$.

The backbone is fully frozen, so adding the head does not alter the agent's general behaviour and requires no retraining or fine-tuning of the backbone itself; we train the head from cached features, with the full training configuration in Appendix~\ref{app:training-config}.


\FloatBarrier
\section{Experiments}
\label{sec:experiments}

\begin{table}[!t]
\centering
\caption{End-to-end quality and token consumption on the read-only multi-turn benchmarks.}
\label{tab:extra-baselines}
\begin{threeparttable}
\footnotesize
\setlength{\tabcolsep}{4.8pt}
\renewcommand{\arraystretch}{1.12}
\begin{tabular*}{\textwidth}{@{\extracolsep{\fill}}lcccccc@{}}
\toprule
& \multicolumn{2}{c}{\textbf{SWE-QA}}
& \multicolumn{2}{c}{\textbf{SWE-QA-Pro}}
& \multicolumn{2}{c}{\textbf{Oolong}} \\
\cmidrule(lr){2-3}\cmidrule(lr){4-5}\cmidrule(lr){6-7}
\rowcolor{headerbg}
\textbf{Method} & \textbf{Score} & \textbf{Tokens}
& \textbf{Score} & \textbf{Tokens}
& \textbf{Acc.} & \textbf{Tokens} \\
\midrule
\multicolumn{7}{l}{\textit{Qwen3-Coder-Next}} \\
\addlinespace[1pt]
No Pruning                  & 7.71 & 590K & 7.60 & 607K & 81.7 & 3.6K \\
\cmidrule(lr){1-7}
LLMLingua2                  & 7.06\loss{0.65} & \textbf{363K}\reduce{38.5\%} & 7.02\loss{0.58} & 381K\reduce{37.3\%} & 74.6\loss{7.1} & 9.5K\worsen{163.9\%} \\
Selective Context           & 7.57\loss{0.14} & 449K\reduce{23.9\%} & 7.32\loss{0.28} & 463K\reduce{23.7\%} & \textbf{81.0}\loss{0.7} & 12.0K\worsen{233.3\%} \\
RAG                         & \textbf{7.78}\improve{0.07} & 549K\reduce{6.9\%} & 7.73\improve{0.13} & 590K\reduce{2.8\%} & 79.1\loss{2.6} & 4.5K\worsen{25.0\%} \\
Self-Prune                  & 7.16\loss{0.55} & 496K\reduce{15.9\%} & 6.96\loss{0.64} & 536K\reduce{11.7\%} & 79.6\loss{2.1} & 3.8K\worsen{5.6\%} \\
LongCodeZip                 & 7.45\loss{0.26} & 531K\reduce{10.0\%} & 7.34\loss{0.26} & 593K\reduce{2.3\%} & 80.1\loss{1.6} & 3.7K\worsen{2.8\%} \\
SWE-Pruner                  & 7.33\loss{0.38} & 397K\reduce{32.7\%} & 7.36\loss{0.24} & 433K\reduce{28.7\%} & 79.7\loss{2.0} & 3.6K \\
\rowcolor{oursbg}
\approach                   & 7.73\improve{0.02} & \textbf{385K}\reduce{34.7\%} & \textbf{7.84}\improve{0.24} & \textbf{368K}\reduce{39.4\%} & 80.3\loss{1.4} & \textbf{3.1K}\reduce{13.9\%} \\
\addlinespace[5pt]
\multicolumn{7}{l}{\textit{MiMo-V2-Flash}} \\
\addlinespace[1pt]
No Pruning                  & 8.02 & 321K & 7.97 & 438K & 92.4 & 58.9K \\
\cmidrule(lr){1-7}
LLMLingua2                  & 7.73\loss{0.29} & 422K\worsen{31.5\%} & 7.47\loss{0.50} & 423K\reduce{3.4\%} & 87.1\loss{5.3} & 170.7K\worsen{189.8\%} \\
Selective Context           & 8.06\improve{0.04} & 360K\worsen{12.1\%} & 7.94\loss{0.03} & 493K\worsen{12.6\%} & 91.8\loss{0.6} & 47.5K\reduce{19.4\%} \\
RAG                         & 8.05\improve{0.03} & 326K\worsen{1.6\%} & \textbf{7.97} & 482K\worsen{10.0\%} & 92.5\improve{0.1} & 47.9K\reduce{18.7\%} \\
Self-Prune                  & 7.75\loss{0.27} & 330K\worsen{2.8\%} & 7.29\loss{0.68} & 466K\worsen{6.4\%} & 91.1\loss{1.3} & 48.6K\reduce{17.5\%} \\
LongCodeZip                 & 8.13\improve{0.11} & 364K\worsen{13.4\%} & 7.93\loss{0.04} & 488K\worsen{11.4\%} & 90.7\loss{1.7} & 46.8K\reduce{20.5\%} \\
SWE-Pruner                  & \textbf{8.20}\improve{0.18} & 303K\reduce{5.6\%} & 7.94\loss{0.03} & 417K\reduce{4.8\%} & 92.1\loss{0.3} & 52.5K\reduce{10.9\%} \\

\rowcolor{oursbg}
\approach                   & 7.98\loss{0.04} & \textbf{299K}\reduce{6.9\%} & 7.86\loss{0.11} & \textbf{339K}\reduce{22.6\%} & \textbf{94.6}\improve{2.2} & \textbf{41.2K}\reduce{30.1\%} \\
\bottomrule
\end{tabular*}
\begin{tablenotes}[flushleft]
\footnotesize
\item Arrows report changes relative to the corresponding \textit{No Pruning} baseline. Green denotes a favorable change and red an unfavorable change. Bold values are best among pruning methods; shaded rows denote \approach.
\end{tablenotes}
\end{threeparttable}
\end{table}

\paragraph{Benchmarks.}
We evaluate on four benchmarks: \emph{SWE-Bench Verified}~\citep{jimenez2024swe} (500 patch-generation issues), \emph{SWE-QA}~\citep{peng2026sweqa} (144 multi-turn QA questions), \emph{SWE-QA-Pro}~\citep{cai2026sweqapro} (260 QA questions with executable environments), and \emph{Oolong}~\citep{oolong} (280 long-context aggregation instances, run as a multi-turn agent variant; Appendix~\ref{app:oolong}). SWE-Bench Verified uses the standard Mini-SWE-Agent harness;\footnote{\url{https://github.com/SWE-agent/mini-swe-agent}} the other three use a minimal bash-only agent (Table~\ref{tab:extra-baselines}). \approach prunes tool outputs throughout.

\paragraph{Agent backbones.} We use two open-weight Mixture-of-Experts LLMs designed for long-horizon coding-agent workloads. \textsc{MiMo-V2-Flash}~\citep{mimo} is a 309B-parameter MoE (15B active per token) with a hybrid sliding-window plus global attention architecture and a 256K context length. \textsc{Qwen3-Coder-Next}~\citep{qwen3codernext} is a more compact 80B-parameter MoE (3B active) built on Qwen3-Next, agent-specialized via large-scale execution-feedback training, with the same 256K context length. Both backbones are served on our patched SGLang stack (Appendix~\ref{app:sglang-patches}).

\paragraph{Baselines.} The unpruned agent (\emph{No Pruning}) is the headline reference. For pruner-vs-pruner comparison on the SWE-QA family and Oolong (Table~\ref{tab:extra-baselines}), we run six prior pruners under the matched configuration, swapping only the pruning module: \emph{LLMLingua2}~\citep{pan2024llmlingua}, \emph{Selective Context}~\citep{li2023compressing}, \emph{RAG} (sliding-window retrieval with bge-reranker-v2-m3\footnote{\url{https://huggingface.co/BAAI/bge-reranker-v2-m3}}), \emph{Self-Prune} (the same agent backbone re-prompted on each $r_t$ with our line-keep labelling prompt), \emph{LongCodeZip}~\citep{shi2025longcodezip} (coarse-only function-level perplexity ranking), and \emph{SWE-Pruner}~\citep{swepruner2026}, the closest prior task-specific pruner.

\paragraph{Metrics and judges.} For SWE-Bench Verified we report the standard Resolve Rate~\citep{jimenez2024swe}. For SWE-QA and SWE-QA-Pro we report mean LLM-judge score on a $1$--$10$ rubric, using \textsc{GPT-5.4-mini} as the judge. For Oolong we use the rule-based exact-match scorer and report accuracy on a $0$--$100$ scale. The annotation and judge prompts are listed in Appendix~\ref{app:prompts}. Across all benchmarks we additionally report total prompt and completion tokens aggregated across all instances, and, where applicable, the number of pruning invocations triggered.

\paragraph{Inference configuration.} Each backbone decodes under the official settings recommended on its HuggingFace model card. Agent rollouts use a maximum of $250$ turns per trajectory; trajectories exceeding this are terminated. The judge \textsc{GPT-5.4-mini} runs at temperature $0$ for determinism. All pruners on the same backbone share identical decoding, harness, and hardware, so pruning is the only varying factor; full serving details are in Appendix~\ref{app:sglang-patches}.

\FloatBarrier
\section{Results}
\label{sec:results}

\subsection{Main Results}
\label{subsec:main-results}

\paragraph{Read-only multi-turn benchmarks.}
Pruning is only worthwhile if it actually shrinks the agent's token bill end-to-end. On SWE-QA, SWE-QA-Pro, and Oolong (Table~\ref{tab:extra-baselines}), \approach is the only pruner that reduces tokens on every cell, with savings up to $39\%$ on the heaviest cell (Qwen3-Coder-Next, SWE-QA-Pro) and $30\%$ on the longest-context one (MiMo-V2-Flash, Oolong). Four of six prior pruners inflate tokens on at least one cell, with LLMLingua2 reaching $+190\%$ on Oolong. The takeaway is that the pruning signal is already in the backbone's representations, so the overhead other pruners pay (goal-hint queries, surrogate scorers, or naive sliding-window retrieval) wipes out their compression gains; reading the signal in place avoids that overhead by construction.

Token savings would be hollow if quality collapsed. With Qwen3-Coder-Next, every pruner except RAG degrades judge scores by $0.14$--$0.65$ on SWE-QA/Pro; RAG preserves quality but yields at most $6.9\%$ token savings—negligible given the overhead it introduces. \approach is the only method that simultaneously maintains quality ($+0.02$, $+0.24$, $-1.4$ pp) and delivers large reductions ($34.7\%$, $39.4\%$, $13.9\%$). The pattern holds on MiMo-V2-Flash: methods that preserve scores (RAG, Selective Context) inflate tokens on four of six cells, while those that compress (SWE-Pruner, Self-Prune) concede quality elsewhere; \approach again achieves the lowest token count on every cell, with its only loss a marginal $-0.11$ on SWE-QA-Pro and a $+2.2$ pp gain on Oolong. Across both backbones, no prior pruner resolves this quality–compression trade-off: the signal already in the backbone is sharp enough to drive aggressive pruning without discarding what the agent needs. Appendix~\ref{app:cases} illustrates this on four representative tool outputs.
\begin{table}[!t]
\centering
\caption{SWE-Bench Verified results across two agent backbones.}
\label{tab:swebench}
\begin{threeparttable}
\small
\setlength{\tabcolsep}{8pt}
\renewcommand{\arraystretch}{1.12}
\begin{tabular}{@{}lccc@{}}
\toprule
\textbf{Method} & \textbf{Resolved} & \textbf{Input Tokens} & \textbf{API Calls} \\
\midrule

\rowcolor{headerbg}
\multicolumn{4}{c}{\textit{MiMo-V2-Flash}} \\

No Pruning
  & 326/500
  & 2,971K
  & 94.8 \\
\cmidrule(lr){1-4}
LongCodeZip
  & 344/500\improve{3.6\%}
  & \textbf{3,166K}\worsen{6.6\%}
  & \textbf{99.8} \\
RAG
  & 338/500\improve{2.4\%}
  & 3,391K\worsen{14.1\%}
  & 102.4 \\
SWE-Pruner
  & \textbf{347/500}\improve{4.2\%}
  & 3,414K\worsen{14.9\%}
  & 103.8 \\
\rowcolor{oursbg}
\approach
  & 345/500\improve{3.8\%}
  & 3,190K\worsen{7.4\%}
  & 111.8 \\

\addlinespace[3pt]

\rowcolor{headerbg}
\multicolumn{4}{c}{\textit{Qwen3-Coder-Next}} \\

No Pruning
  & 341/500
  & 5,307K
  & 131.9 \\
\cmidrule(lr){1-4}
LongCodeZip
  & 288/500\loss{10.6\%}
  & 4,718K\reduce{11.1\%}
  & \textbf{124.7} \\
RAG
  & 330/500\loss{2.2\%}
  & 4,805K\reduce{9.5\%}
  & 126.0 \\
SWE-Pruner
  & 320/500\loss{4.2\%}
  & 4,881K\reduce{8.0\%}
  & 127.1 \\
\rowcolor{oursbg}
\approach
  & \textbf{335/500}\loss{1.2\%}
  & \textbf{4,590K}\reduce{13.5\%}
  & 139.8 \\

\bottomrule
\end{tabular}
\begin{tablenotes}[flushleft]
\footnotesize
\item Input tokens are per-trajectory averages. Arrows report changes from
the same-backbone \textit{No Pruning} baseline. Bold values are best among
pruning methods; shaded rows denote \approach.
\end{tablenotes}
\end{threeparttable}
\end{table}

\paragraph{Code-modification benchmark.}
We turn next to SWE-Bench Verified, which extends the comparison to the heavier patch-generation workload (Table~\ref{tab:swebench}). The picture is asymmetric across backbones. On \textsc{MiMo-V2-Flash}, all pruners improve the resolve rate; \approach reaches $+3.8\%$ at roughly half the token overhead of the next-best pruner (SWE-Pruner: $+4.2\%$ resolve at $+14.9\%$ tokens). On \textsc{Qwen3-Coder-Next}, every pruner instead loses resolves, but \approach gives the most favorable degradation profile, losing only 6 solves (-1.2 points) while achieving the largest input-token reduction (-13.5\%),  outperforming the other pruners on both resolve rate and input-token use. 

API calls provide a separate efficiency signal from input tokens, because pruning changes the history the agent conditions on in later turns rather than only shortening a fixed trajectory. On \textsc{MiMo-V2-Flash}, all pruners increase calls relative to \emph{No Pruning}, with \approach using the most. On \textsc{Qwen3-Coder-Next}, LongCodeZip, RAG, and SWE-Pruner reduce calls, whereas \approach increases calls from 131.9 to 139.8 while still achieving the largest input-token reduction. Because input-token cost and API-call count trade off in opposite directions across backbones, we report them separately rather than collapsing them into a single efficiency number.





\FloatBarrier
\subsection{Ablations}
\label{subsec:ablation}

To quantify how much each design choice contributes, we ablate the loss function and the length-aware embedding while holding everything else fixed (same 22k feature set from \textsc{Qwen3-Coder-Next}). We score the resulting checkpoints on a held-out judge set ($n{=}100$, \textsc{GPT-5.4-mini} on a 1--10 rubric) alongside per-line F1.

\begin{table}[H]
\centering
\caption{Ablation results on the held-out judge set.}
\label{tab:ablation}
\begin{threeparttable}
\small
\renewcommand{\arraystretch}{1.12}
\begin{tabular}{@{}llcc@{}}
\toprule
\textbf{Design axis} & \textbf{Variant} & \textbf{$F_1$} & \textbf{Judge} \\
\midrule
\multirow{5}{*}{\textit{Loss function}}
  & BCE & 0.475 & 5.95 \\
  & Focal & 0.593 & 6.37 \\
  & Dice & 0.591 & 5.30 \\
  & Tversky & 0.591 & 3.03 \\
  & \cellcolor{oursbg}Per-sample balanced focal $\star$
  & \cellcolor{oursbg}\textbf{0.635}
  & \cellcolor{oursbg}\textbf{7.08} \\
\addlinespace[3pt]
\midrule
\multirow{2}{*}{\textit{Length-aware embedding}}
  & Without length embedding & \textbf{0.636} & 6.86 \\
  & \cellcolor{oursbg}With length embedding $\star$
  & \cellcolor{oursbg}0.635
  & \cellcolor{oursbg}\textbf{7.08} \\
\bottomrule
\end{tabular}
\begin{tablenotes}[flushleft]
\footnotesize
\item Each design axis is varied while the other is held at its default ($\star$).
Judge: \textsc{GPT-5.4-mini}; $F_1$ is measured per line.
\end{tablenotes}
\end{threeparttable}
\end{table}

\paragraph{Loss function.}
We compare per-sample balanced focal against four alternatives: BCE, corpus-level Focal, Dice, and Tversky (Table~\ref{tab:ablation}, loss-function block; full hyperparameters in Appendix~\ref{app:training-config}). Per-sample balanced focal wins on both metrics, beating BCE by $+1.13$ judge and $+0.16$ F1. Dice and Tversky match its F1 but their judge collapses, because per-line label match cannot tell a useful skeleton from a precision-correct but unusable one (Appendix~\ref{app:f1-vs-judge}). The takeaway is that the keep-or-prune imbalance is per-sample rather than global, and rebalancing inside each sample is what unlocks the gain.

\paragraph{Length-aware embedding.}
Adding the length-aware embedding lifts judge from $6.86$ to $7.08$ at essentially identical F1 (Table~\ref{tab:ablation}, length-aware-embedding block). The embedding does not change overall line-decision accuracy; it redistributes mistakes toward longer responses where mis-pruning a single line is far less harmful, exactly the length-conditioned asymmetry Eq.~\ref{eq:size-bias} is designed to capture.

\subsection{Latency analysis}
\label{subsec:latency}

\begin{figure}[H]
\centering
\includegraphics[
  max width=0.82\linewidth,
  max totalheight=0.31\textheight
]{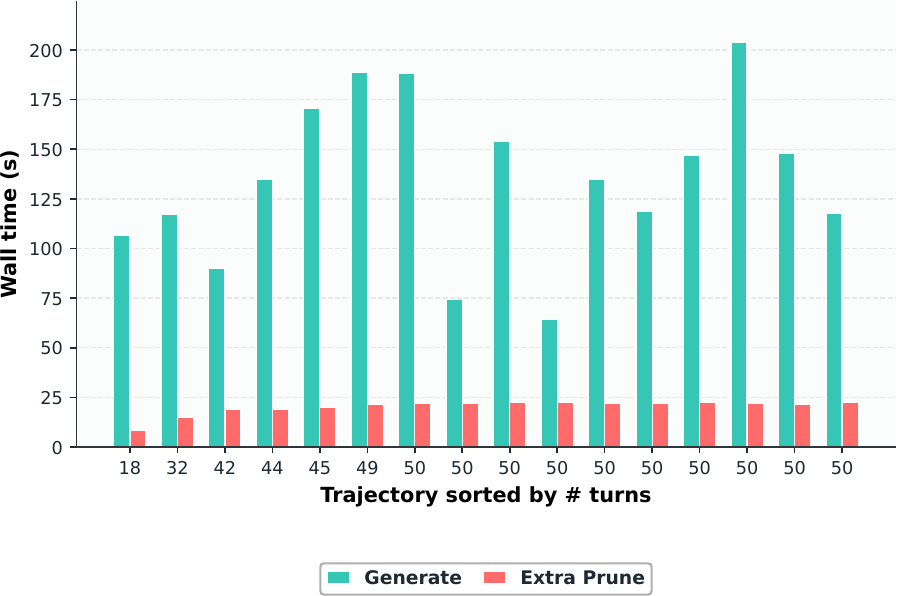}
\caption{Per-call pruning overhead relative to the immediately following generation step, broken down by trajectory.}
\label{fig:latency}
\end{figure}

On a 144-trajectory \textsc{MiMo-V2-Flash} SWE-QA replay (setup details in Appendix~\ref{app:hs-extraction}), with the in-engine pruning head of Appendix~\ref{subsec:in-engine} enabled, the pruning calls add $8.4\%$ aggregate wall time relative to total generation time, with per-trajectory ratios at $p_{50}=9.1\%$ and $p_{95}=19.9\%$. Figure~\ref{fig:latency} illustrates generation and pruning times for individual trajectories. The bound holds because the head reuses the prefill the backbone already runs on the new tool response, and colocating the head inside the inference engine avoids transferring hidden states across the engine boundary, so the only added compute is a single head forward. This per-call overhead is paid roughly once per assistant turn, while the token savings reported in Table~\ref{tab:extra-baselines} reduce backbone work on every subsequent turn, so the net wall-clock impact across a trajectory is dominated by the token reduction rather than the pruning overhead. The overhead profile is expected to shrink further in heavier-decode regimes and with tighter hidden-state quantisation; we discuss the off-engine variant and remaining headroom in Appendix~\ref{subsec:in-engine}.

\FloatBarrier
\section{Related Work}
\label{sec:related}

\subsection{Prompt Compression for Code}
\label{subsec:related-compression}
Reducing the prompt length seen by a code-oriented LLM~\citep{rando2025longcodebench}, now widely applied from code generation~\citep{shi2024code,chen2026classeval} to repository-level issue resolution~\citep{jimenez2024swe}, has been approached along three broad axes. Token-level pruners score tokens by self-information or perplexity and drop the lowest-ranking ones~\citep{jiang2023llmlingua,jiang2024longllmlingua,li2023compressing,fang2025attentionrag}; retrieval-based shorteners replace verbatim content with retrieved or aggregated snippets~\citep{lewis2020retrieval,cheng2024xrag,cheng2026resolving,shi2026reasoning,zhang2023repocoder,lai2026transformers}. Both families optimise content-level signals only and disregard the syntactic and structural constraints that code understanding requires~\citep{shi2025longcodezip,shi2025between}. A third, code-aware family preserves program structure during compression~\citep{zhang2022diet,wang2024natural,shi2025longcodezip,shi2026codeocr,hu2026line,zeng2025pruning}, but is largely evaluated on single-round proxy tasks rather than multi-turn agent loops, and applies fixed policies~\citep{yang2024less} regardless of the agent's intent. The closest work targeting coding agents end-to-end is \mbox{SWE-Pruner}~\citep{swepruner2026}, which performs line-level pruning conditioned on an explicit goal-hint query at each turn; \approach inherits its line-level granularity and benchmark setting but discards both the separate scoring model and the explicit query, reading the signal directly from the agent backbone's hidden states.

\subsection{Agent Context Management}
\label{subsec:related-agent}
Even with the 128k-plus context windows of current coding LLMs~\citep{achiam2023gpt,team2024gemini}, agents on real codebases routinely overflow them and the underlying models suffer quality drops as context grows~\citep{liu2023lost,laban2025llms,li2023loogle}, making efficiency a first-class engineering problem for software-engineering agents~\citep{yao2023react,wei2022chain,yang2024sweagent,wang2024openhands,xia2024agentless,bouzenia2024repairagent,qin2024agentfl,liu2024large,fang2025lastingbench,zhao2026dllm,chen2025swe,li2025swe,zhang2026swe,DBLP:journals/corr/abs-2609-02783,shi2026speculativeforkingturningidle}. Practical deployments cope by summarising the trajectory on overflow, hard-truncating the prefix, or masking older observations~\citep{cursor,claude_code,gao2025trae,lindenbauer2025complexity}; recent learned approaches train policies that manage history through reinforcement learning, hierarchical oversight, or proactive folding~\citep{lu2025supo,sun2025contextfolding,wan2025compass,kang2025acon,xiao2026agentdiet,ye2025agentfold,gao2026swe,zhang2026fastcontext}.
All of these operate on the agent's prior interaction history; \approach instead operates one level upstream, at the agent-environment boundary, by compressing incoming environment observations before they enter the history.
The same boundary is targeted by SWE-Pruner~\citep{swepruner2026}; where it reconstructs the pruning signal with a dedicated external model, SWE-Pruner Pro instead reads it from representations the backbone has already formed.

\FloatBarrier
\section{Conclusion}
\label{sec:conclusion}
We have shown that the internal representations a coding agent forms while reading a tool output already encode line-level importance, removing the need for a separate scoring model or an explicit goal-hint query. \approach reads this signal out through a lightweight head that shares the backbone's prefill pass. Two design choices carry most of the empirical lift, namely a learned length-aware embedding and a per-sample balanced focal loss. The result is consistent across two open-weight backbones and four multi-turn benchmarks, where \approach reduces token consumption while keeping quality within a narrow band of the unpruned baseline, at negligible added inference cost. This points to a broader principle: the representations a backbone forms while passively reading an observation already encode the relevance judgment that prior approaches spent an extra model call to recover. Rather than building pruning around the agent, it suffices to read the signal the agent has already formed.

\section*{Limitations}
\label{sec:limitations}

We close with two scope notes. \approach reads the agent backbone's internal hidden states, so the present evaluation covers only open-weight models; the recipe itself extends to any backbone that exposes its hidden states, and the consistent quality preservation across the SWE-QA family and the natural-language Oolong setting suggests the paradigm transfers across both code and natural-language tool outputs, with only per-backbone head retraining required for a new model. As for language coverage, although our agent-task benchmarks are Python-centric, the pipeline is language-agnostic, and Oolong (a long-context natural-language aggregation benchmark) gives an out-of-domain check on which \approach preserves both token savings and quality on both backbones (Table~\ref{tab:extra-baselines}); broader coverage across programming languages reuses the same pipeline and is left to future work.


\section*{Ethical Considerations}
All training trajectories are drawn from publicly released datasets; no private repositories or proprietary codebases were used. The pruning head is trained solely to compress redundant tool output and does not alter the agent's generation or reasoning; deployment should be validated per-backbone before use in safety-critical settings, as aggressive pruning may degrade task quality in ways not captured by our benchmarks.

\FloatBarrier
\bibliographystyle{plainnat}
\bibliography{custom}

\clearpage
\appendix

\section{Training Data}
\label{app:training-data}

\begin{figure}[H]
\centering
\includegraphics[
  max width=0.84\linewidth,
  max totalheight=0.24\textheight
]{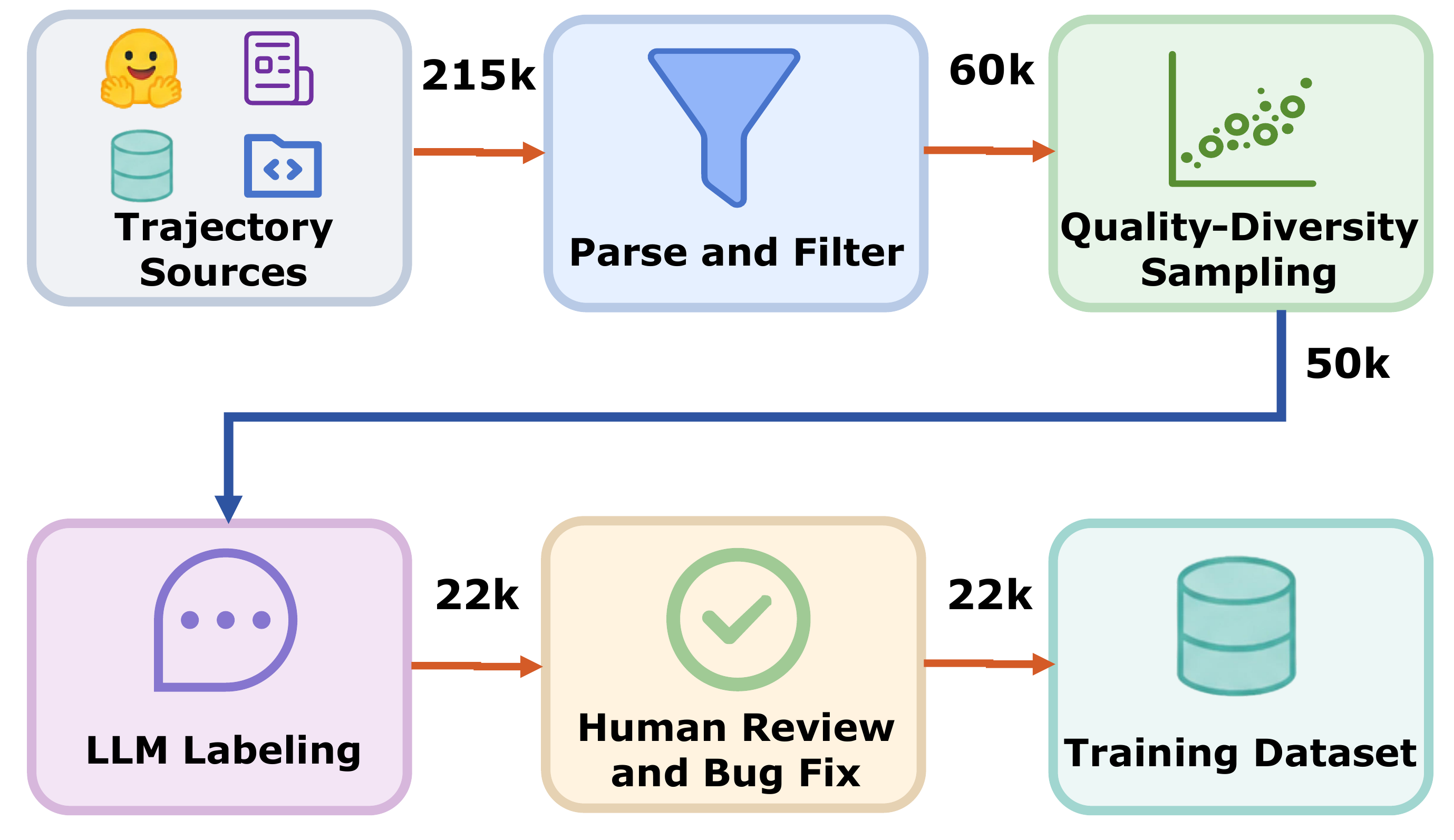}
\caption{Data construction pipeline.
Raw trajectories from five HuggingFace sources are parsed into a uniform per-step schema, light-filtered, then sub-sampled to a 50k diverse pool by quality-diversity selection, labelled per line by \textsc{Claude Sonnet~4.6}, and finally passed through a small human review pass that drops a few hundred of malformed labels.}
\label{fig:data-pipeline}
\end{figure}

\paragraph{Overview.}
The training corpus aggregates agent trajectories from five publicly released HuggingFace datasets: four are SWE-style code-modification rollouts and one (\texttt{terminal-wrench-trajectories}) covers a mix of CLI / shell tasks spanning chess, machine learning, cryptography, databases and shell scheduling.
We deliberately include both code-patch and non-code agent traces so that the \approach head learns to handle the heterogeneous tool outputs encountered in practice.
After all stages of the pipeline (Figure~\ref{fig:data-pipeline}) the labelled corpus contains $22{,}609$ $(history,\ tool\_call,\ tool\_response)$ samples drawn from $6{,}252$ unique trajectories (${\approx}3.6$ steps per trajectory on average); the per-source breakdown is given in Table~\ref{tab:training-data-sources}.

\begin{table}[H]
\centering
\caption{Source-level composition of the labelled training corpus.}
\label{tab:training-data-sources}
\begin{threeparttable}
\small
\setlength{\tabcolsep}{7pt}
\renewcommand{\arraystretch}{1.10}
\begin{tabular}{@{}lrrr@{}}
\toprule
\textbf{HuggingFace dataset}
  & \textbf{Traj.}
  & \textbf{Samples}
  & \textbf{Share} \\
\midrule
\texttt{few-sh/terminal-wrench-trajectories}~\citep{bercovich2026terminalwrenchdataset331}
  & 3\,349 & 6\,848 & 30.3\% \\
\texttt{TIGER-Lab/SWE-Next-SFT-Trajectories}~\citep{liang2026swenextscalablerealworldsoftware}
  & 899 & 6\,632 & 29.3\% \\
\texttt{ByteDance-Seed/Multi-SWE-bench\_trajs}~\citep{zan2025multiswebench}
  & 872 & 3\,760 & 16.6\% \\
\texttt{zai-org/CC-Bench-trajectories}~\citep{zai2025ccbenchtrajectories}
  & 280 & 3\,074 & 13.6\% \\
\texttt{AweAI-Team/Scale-SWE-Distilled}~\citep{zhao2026immersiongithubuniversescaling}
  & 852 & 2\,295 & 10.2\% \\
\midrule
\textbf{Total}
  & \textbf{6\,252}
  & \textbf{22\,609}
  & \textbf{100\%} \\
\bottomrule
\end{tabular}
\begin{tablenotes}[flushleft]
\footnotesize
\item A sample is one $(\textit{history}, \textit{tool call}, \textit{tool response})$
step; a trajectory is one globally deduplicated \texttt{instance\_id}.
\end{tablenotes}
\end{threeparttable}
\end{table}

\paragraph{Pre-filtering and diversity sampling.}
A per-source adaptor parses each raw trajectory into a flat list of chat messages following the OpenAI tool-calling schema (\texttt{system / user / assistant / tool} roles), and emits one $(history,\ tool\_call,\ tool\_response,\ next\_turn)$ quadruple for every \texttt{tool} message.
The \texttt{history} is kept as a sliding window that preserves complete \texttt{(assistant, tool)} turn pairs; partial pairs that do not fit are dropped rather than truncated, so \texttt{tool\_calls} are never split.
A light pre-filter drops responses with fewer than two non-blank lines, implausibly low code-line ratios, and pure stack-trace error output, so that the labelling budget is spent on tool responses with substantive content.
The surviving pool is larger than what we want to label, so we sub-sample it to a 50k diverse subset. The selection maximises a facility-location objective over instance metadata (language, repo, response-size bucket and a log-step length kernel) under a per-instance step cap.
This is run greedily with lazy evaluation, which gives the standard $(1{-}1/e)$ approximation guarantee and keeps the selection cost negligible relative to labelling.

\paragraph{Per-line labelling.}
Each retained quadruple is labelled by \textsc{Claude Sonnet~4.6} via the Anthropic API.
The labeller sees the full \texttt{history}, the triggering \texttt{tool\_call}, the line-numbered \texttt{tool\_response} (with internal \texttt{cat -n} numbers stripped to avoid confusing inner with outer indices), and the \texttt{next\_turn} snippet. It emits a list of 1-based line indices that the agent must retain, together with a short \texttt{reasoning} field and a \texttt{confidence} flag.
About half of the input pool yields a non-empty label set; the rest is naturally rejected by the labeller as containing nothing worth retaining (boilerplate, pure logs, etc.) and dropped.
A separate category covers responses where the labeller cannot identify specific lines to prune. These are marked \texttt{skeleton} when the labeller declares the entire output is a scaffold the agent will reuse downstream (${\approx}17\%$ of the corpus, mostly in TIGER-Lab, CC-Bench and Multi-SWE-bench); for these samples every token in the tool response receives label $1$, giving the head explicit examples of the ``do not prune'' regime.
At training time, the kept-line set is expanded to per-token binary labels (a token receives label $1$ if its character span overlaps at least one kept line) and tokens outside the \texttt{<tool\_response>{\ldots}</tool\_response>} span are masked out of the loss.
A final manual-review pass removes a few hundred labels with obvious inner/outer line-number confusion, empty kept-sets on non-trivial responses, and malformed \texttt{instance\_id}s, leaving the $22{,}609$-sample corpus used throughout the paper.

\paragraph{Distribution.}
The \texttt{tool\_response} length distribution is heavy-tailed but bounded by an upstream length cutoff: mean $76$ lines, median $56$ lines, with the $10/90/99$ percentiles at $25/143/294$ and a hard maximum of $465$ lines.
The \textsc{Claude}-assigned keep-ratio (kept lines $\div$ total lines) has mean $0.32$ and median $0.23$, matching our compression target of ${\approx}30\%$.
In terms of language coverage, languages are recovered from \texttt{instance\_id} for \texttt{Multi-SWE-bench\_trajs} (which encodes the language as \texttt{mswe\_<lang>\_\ldots}) and inferred from the source dataset for the others: \texttt{Scale-SWE-Distilled} and \texttt{TIGER-Lab/SWE-Next} are Python-only, \texttt{CC-Bench} is bash-dominant with some Python, and \texttt{terminal-wrench} is a domain-mixed CLI corpus. The resulting breakdown is shown in Table~\ref{tab:training-data-lang}, grouped into single-language rows (top) and mixed/CLI rows (bottom). Python and CLI/shell-style content together account for ${\approx}83\%$ of the corpus; six explicit non-Python languages from \texttt{Multi-SWE-bench\_trajs} make up the remaining ${\approx}17\%$.

\begin{table}[!t]
\centering
\caption{Language and category composition of the training corpus.}
\label{tab:training-data-lang}
\small
\setlength{\tabcolsep}{10pt}
\renewcommand{\arraystretch}{1.10}
\begin{tabular}{@{}lrr@{}}
\toprule
\textbf{Language / category} & \textbf{Samples} & \textbf{Share} \\
\midrule
Python                & 8\,927 & 39.5\% \\
C++                   &   976 &  4.3\% \\
JavaScript            &   828 &  3.7\% \\
C                     &   709 &  3.1\% \\
Java                  &   561 &  2.5\% \\
TypeScript            &   500 &  2.2\% \\
Go                    &   186 &  0.8\% \\
\addlinespace[4pt]
Bash (dominant)       & 3\,074 & 13.6\% \\
Domain-mixed (CLI)    & 6\,848 & 30.3\% \\
\bottomrule
\end{tabular}
\end{table}

\paragraph{Probe subset for \S\ref{sec:motivation}.}
The motivation analysis in \S\ref{sec:motivation} uses a random $10\%$ subset of the labelled corpus drawn at the trajectory level ($\approx 625$ trajectories, $\approx 2{,}260$ tool responses, $\approx 155$k lines). Within this subset we further split trajectories $90/10$ into train/eval for the linear probe, so no trajectory contributes lines to both splits; the AUC $0.83$ and best-$F_1$ $0.63$ reported in Figure~\ref{fig:linear-probe} are computed on the held-out trajectories.
The subset is intentional: the analysis is an existence proof for the keep-or-prune signal in the backbone's hidden states, and a logistic regression on a fraction of the corpus is sufficient to test separability.
The head architecture and training in \S\ref{sec:method} use the full corpus.


\FloatBarrier
\section{Training Details}
\label{app:training-config}

\paragraph{Head architecture.}
The pruning head is a per-token feed-forward MLP applied to the frozen backbone's last-layer hidden states $h_i$. The continuous length-aware embedding $\mathbf{e}(N)$ of Eq.~\ref{eq:size-bias} is realised over $8$ log-spaced \texttt{n\_lines} buckets (0--2, 3--5, 6--10, 11--20, 21--50, 51--100, 101--200, ${>}200$), additively combined with each hidden state in the tool-response span and zero-initialised so $f_\theta$ reduces to its length-agnostic limit at the start of training. The classifier $f_\theta$ itself is a LayerNorm followed by two \texttt{Linear-GELU-Dropout} blocks with hidden dimension matching the backbone hidden size $d$ (so the head is re-sized per backbone) and dropout $0.4$, then a final Linear projection to a single keep logit.

\paragraph{Optimisation.}
The head is optimised with AdamW ($\beta_1{=}0.9$, $\beta_2{=}0.999$, $\epsilon{=}10^{-8}$, weight decay $0$) at a peak learning rate of $3\times10^{-5}$.
The schedule is a linear warmup over the first $5\%$ of total updates, followed by cosine decay to a floor of $1.5\times10^{-5}$, so the effective learning rate stays in $[1.5\times10^{-5},\,3\times10^{-5}]$ throughout the run.
Gradient clipping with max-norm $1.0$ is applied to the head parameters.
Random seeds are fixed (\texttt{seed=42}); no backbone parameters are updated at any point.

\paragraph{Loss.}
The training objective is the per-sample class-balanced focal loss of Eq.~\ref{eq:loss-sample}, with focal exponent $\gamma=2$ (Eq.~\ref{eq:focal}) and the equal-weight $0.5/0.5$ keep/prune mix of Eq.~\ref{eq:balance}. Samples that happen to contain only one class contribute the surviving branch with weight $1$ and drop the absent branch from the average.

\paragraph{Frozen-backbone, feature-cache pipeline.}
Hidden states are pre-extracted from the frozen backbone once per dataset~$\times$~backbone pair and cached on disk as memory-mapped feature files; all subsequent head training reads directly from these features rather than re-running the backbone.
A full $10$-epoch pass over the $22{,}609$ samples completes in $\approx 15$~min on a single $8\times\text{H200}$ node, which makes the loss / length-bias ablation grid in \S\ref{subsec:ablation} cheap to run end-to-end.

\FloatBarrier
\section{Oolong Benchmark Conversion}
\label{app:oolong}

\paragraph{Oolong as a multi-turn agent benchmark.}
Oolong was originally released as a single-turn benchmark: each of the 280 examples contains a long synthetic context (10K--65K tokens, drawn from \textsc{agnews}, \textsc{imdb}, \textsc{multinli}, \textsc{spam}, etc.) and a counting-style question (e.g., most-common label, label frequency), and a model is expected to produce the answer in a single pass from the full context.
Long-context aggregation is a natural fit for agentic exploration: the model can locate and aggregate the relevant rows itself rather than scanning the full context end-to-end. We therefore re-cast Oolong into a multi-turn tool-use setting that matches the harness of our other benchmarks, and reuse the official scorer.

\paragraph{Multi-turn re-cast.}
For each example, the long context is materialised as a read-only \texttt{data.txt} inside a Docker sandbox; the model is no longer fed the context directly, but is instead given a single \texttt{bash} tool and asked to explore the file with standard CLI utilities (\texttt{grep}, \texttt{awk}, \texttt{sort}, \texttt{uniq~-c}, \texttt{wc}, \ldots) until it produces an \texttt{Answer:} line.
Tool outputs are returned to the model as \texttt{tool\_response} messages and pruned on-the-fly by each method under test, exactly as in our other multi-turn benchmarks.
Scoring reuses Oolong's official \texttt{synth\_process\_response} routine, so numbers remain directly comparable to the original single-turn leaderboard.
In practice this turns each question into a short trajectory (typically 3--14 tool calls, depending on backbone), where the agent must locate and aggregate the relevant rows itself rather than relying on the model to scan the full context end-to-end.

\clearpage
\section{Per-turn execution}
\label{app:perturn}
Algorithm~\ref{alg:perturn} expands the per-turn pipeline of \S\ref{subsec:pipeline} as pseudocode, including the turn-boundary substitution that replaces $r_t$ with the pruned $\tilde r_t$ in the history $H_t$ used by turn $t{+}1$.

\begin{algorithm}[H]
\small
\caption{Per-turn execution of \approach}
\label{alg:perturn}
\begin{algorithmic}[1]
\Require History $H_{t-1}$, tool call $c_t$, raw response $r_t$ with $N$ lines, head $f_\theta$, threshold $\tau$
\Ensure Pruned response $\tilde r_t$
\State $\{h_i\}_{i=1}^{L} \gets \mathrm{Prefill}(r_t \mid H_{t-1}, c_t)$ \Comment{only $r_t$ forwarded; prefix cached}
\State $\mathbf{b} \gets \mathbf{e}(N)$ \Comment{length-aware embedding, Eq.~\ref{eq:size-bias}}
\For{$i = 1, \ldots, L$}
    \State $\tilde h_i \gets h_i + \mathbf{b}$
    \State $p_i \gets \sigma\!\left(f_\theta(\tilde h_i)\right)$ \Comment{Eq.~\ref{eq:logit}}
\EndFor
\For{each line $\ell$}
    \State $\hat y_\ell \gets \mathbb{1}\!\left[\tfrac{1}{|\ell|}\sum_{i\in\ell} \mathbb{1}[p_i > \tau] > \tfrac{1}{2}\right]$ \Comment{Eq.~\ref{eq:majority}}
\EndFor
\State $\tilde r_t \gets r_t$ with all lines $\ell$ satisfying $\hat y_\ell = 0$ removed
\State Substitute $\tilde r_t$ for $r_t$ when forming $H_t$ for turn $t{+}1$
\State \Return $\tilde r_t$
\end{algorithmic}
\end{algorithm}

\section{In-server hidden-state extraction}
\label{app:hs-extraction}

The \approach head consumes per-token last-layer hidden states from the prefill the backbone already performs on each tool response (\S\ref{subsec:pipeline}). We serve the backbone on SGLang\footnote{\url{https://github.com/sgl-project/sglang}} 0.5.10.post1 with \texttt{-{}-enable-return-hidden-states}, which ships those hidden states back inside the generation response. In practice we found three correctness gaps on the hidden-state path where guards already enforced on the logprob path were missing: alignment, chunked-prefill, and prefix-cache. We additionally found that routing hidden states through the existing JSON response field is impractical for tensors three orders of magnitude larger than a typical text payload. We close these gaps below in three stages --- three correctness fixes (\S\ref{subsec:hs-correctness}), a payload-size fix (\S\ref{subsec:hs-payload}), and an in-engine head colocation that removes the cross-process boundary entirely (\S\ref{subsec:in-engine}) --- then validate the corrected hidden states against a pure-transformers reference and quantify the payload reduction.

\subsection{Correctness: closing the hidden-state vs.\ logprob asymmetry}
\label{subsec:hs-correctness}
\label{app:sglang-patches}

We identified three failure modes on the hidden-state path. Each is the same structural problem in different clothing: a guard that the logprob path already enforces is simply absent on the hidden-state side. We restore symmetry rather than introduce new mechanisms, which keeps the patches small, local, and consistent with SGLang's existing design idioms.

\paragraph{Batch alignment.}
\texttt{tokenizer\_manager.py} indexes \texttt{recv\_obj.output\_hidden\_states[i]} assuming the list is aligned with \texttt{rids}. Upstream, the scheduler appended to \texttt{output\_hidden\_states} only for requests with \texttt{return\_hidden\_states=True}, so in a mixed batch (agent generation calls interleaved with pruner hidden-state requests) the list ran shorter than the batch and the index raised \texttt{IndexError}. The logprob path already preallocates a per-request slot in this situation. We mirror that on the hidden-state side: preallocate \texttt{[] if any(r.return\_hidden\_states for r in reqs) else None} in \texttt{stream\_output\_generation} and always append either the per-request hidden states or \texttt{None}, so \texttt{len(output\_hidden\_states) == len(rids)} holds in every batch.

\paragraph{Chunked-prefill accumulation.}
Under chunked prefill each forward emits hidden states only for the current chunk, but the original code only captured them on \texttt{is\_chunked<=0} (the final chunk), so for any prompt that crossed the chunk boundary the returned tensor covered only the tail span and was tagged with the wrong absolute positions. The logprob path already accumulates per-chunk results into the request slot using the per-chunk span recorded in \texttt{extend\_input\_len\_per\_req[i]}; we apply the same accumulation pattern, appending each chunk's hidden states into \texttt{req.hidden\_states[0]} via \texttt{np.concatenate}. Because \texttt{event\_loop\_overlap} overwrites \texttt{req.extend\_input\_len} in place before the consumer reads it, we further widen the existing logprob snapshot guard in \texttt{scheduler.py} from \texttt{batch.return\_logprob} to \texttt{batch.return\_logprob or batch.return\_hidden\_states} so the per-request span is captured for hidden-state requests too. After the patch, \texttt{req.hidden\_states[0]} covers positions $[0, L)$ for any chunked prefill of length $L$.

\paragraph{Prefix-cache exemption.}
SGLang's radix prefix cache stores the KV produced by a previous forward but not the hidden states. When a later request shares a leading token sub-sequence with a cached entry, those positions are skipped during prefill: the KV is reused directly and the model never produces hidden states for them. The returned hidden-state tensor therefore covers only the suffix \texttt{[prefix\_len{:}\,total\_len]}, and any consumer that needs hidden states for a region the cache happens to cover sees a silently truncated tensor. The condition triggers whenever the cached prefix covers tokens whose hidden states the consumer needs, e.g.\ an earlier identical prompt. Logprob has the exact same problem in principle, and SGLang already solves it: \texttt{init\_next\_round\_input} caps \texttt{max\_prefix\_len} at \texttt{logprob\_start\_len} when \texttt{return\_logprob} is on, forcing positions $\geq\texttt{logprob\_start\_len}$ through the forward pass. We mirror this on the hidden-state side by threading a new \texttt{hidden\_states\_start\_len} field through \texttt{GenerateReqInput}, \texttt{TokenizedGenerateReqInput} and \texttt{Req}, and extending the same cap in \texttt{init\_next\_round\_input}:

\begin{lstlisting}[style=codestyle]
if self.return_hidden_states and self.hidden_states_start_len >= 0:
    max_prefix_len = min(max_prefix_len, self.hidden_states_start_len)
\end{lstlisting}

The caller sets \texttt{hidden\_states\_start\_len} to the start of the region it needs hidden states for; tokens before the cap still benefit from cache reuse, while tokens after it are guaranteed to forward and produce hidden states.

\paragraph{Validation against a transformers reference.}
We validate the value-level patches (batch alignment and chunked-prefill accumulation) with a regression harness on a held-out 48-sample probe set bucketed by prompt length ($<\!2$k\,/\,$2$--$8$k\,/\,$8$--$16$k\,/\,$16$--$24$k, 12 samples each), comparing patched SGLang hidden states against a pure-transformers reference path using \texttt{flash\_attention\_2}. The prefix-cache patch cannot be exercised against this reference (transformers has no shared radix cache); we validate it directly by toggling the \texttt{hidden\_states\_start\_len} cap and re-issuing a probe whose prefix is a known superset of an earlier request, then checking that the returned tensor covers the requested region rather than the post-prefix suffix only.

\begin{table}[!t]
\centering
\caption{Validation of patched SGLang hidden states against a pure-transformers \texttt{flash\_attention\_2} reference.}
\label{tab:hs-validation}
\small
\setlength{\tabcolsep}{10pt}
\renewcommand{\arraystretch}{1.10}
\begin{tabular}{@{}lcc@{}}
\toprule
\textbf{Metric} & \textbf{Before patches} & \textbf{After patches} \\
\midrule
Shape match     & 1 / 48 & \textbf{48 / 48} \\
Cosine, median  & n/a    & \textbf{0.997} \\
Cosine, mean    & n/a    & 0.983 \\
\bottomrule
\end{tabular}
\par\vspace{2pt}
{\footnotesize\raggedright Shape match counts samples with matching tensor shape; cosine is computed per token against the reference and then aggregated.\par}
\end{table}

After patching, all 48 samples match shape exactly and per-token cosine reaches $0.997$ at the median (Table~\ref{tab:hs-validation}). The mean is pulled down by a tail with $\cos<0.98$ (min $0.33$), which we attribute to bf16 arithmetic noise from different attention backends and GEMM reduction order rather than to a residual correctness bug.

\subsection{Payload: replacing text-shaped serialization}
\label{subsec:hs-payload}
\label{app:latency-opts}

Because the pruning head reuses the backbone's existing prefill, the only additional traffic over a non-pruning deployment is a single hidden-state tensor crossing the SGLang~$\to$~pruner boundary per tool response. Our initial implementation routed the tensor through SGLang's existing JSON response path: the numpy hidden-state tensor is written into \texttt{meta\_info.hidden\_states} and serialized by \texttt{orjson} as a nested list of fp32 values through \texttt{OPT\_SERIALIZE\_NUMPY}. This default is fine for short text fields but ill-suited to hidden-state tensors: for a 16\,k-token prompt at hidden size 2048 the $16384 \times 2048$ fp32 tensor is 128\,MB of raw bytes but expands to 1--3\,GB after ASCII float formatting, with both the server \texttt{orjson} walk and the client \texttt{json.loads} paying per-element. Two nested optimisations close this gap: replace the nested list with a raw binary envelope to avoid the ASCII-formatting cost, then cast to fp16 to halve the byte count at a controlled precision cost.

We carry the tensor in the same \texttt{meta\_info.hidden\_states} field as a base64 binary envelope:

\begin{lstlisting}[style=codestyle]
arr = recv_obj.output_hidden_states[i][0]
meta_info["hidden_states"] = {
    "__binary__": True, "shape": list(arr.shape),
    "dtype": str(arr.dtype),
    "data": pybase64.b64encode(arr.tobytes()).decode("utf-8"),
}
\end{lstlisting}

\noindent The same envelope carries either fp32 or fp16 depending on \texttt{arr.dtype}; we cast to fp16 on the SGLang side before encoding.

\begin{table}[!t]
\centering
\caption{Per-request hidden-state payload by serialization format.}
\label{tab:hs-payload}
\small
\setlength{\tabcolsep}{12pt}
\renewcommand{\arraystretch}{1.10}
\begin{tabular}{@{}lr@{}}
\toprule
\textbf{Format} & \textbf{Payload size} \\
\midrule
Nested JSON list of fp32 & 1--3\,GB text \\
Binary envelope, fp32    & 170\,MiB \\
Binary envelope, fp16    & \textbf{85\,MiB} \\
\bottomrule
\end{tabular}
\par\vspace{2pt}
{\footnotesize\raggedright Measured for a 16\,k-token prompt with hidden size $H{=}2048$.\par}
\end{table}

The binary envelope removes the ASCII-formatting blowup, and stacking fp16 on top halves the byte count again, giving a roughly $20\times$ end-to-end reduction over the default path (Table~\ref{tab:hs-payload}) at a precision cost bounded by the median-token cosine $0.997$ reported in Table~\ref{tab:hs-validation}. This payload reduction is what keeps the per-call pruning overhead in Figure~\ref{fig:latency} bounded; the default nested-JSON path would dominate per-call wall time on long prompts.

\subsection{In-engine pruning head}
\label{subsec:in-engine}

The patches above describe the off-engine path: hidden states leave the inference engine and the head runs as a separate pruner process. When the head is small (here an $18$M-parameter FFN on top of a multi-billion-parameter MoE backbone), this boundary can be removed by colocating the head inside the SGLang scheduler itself. The head runs directly on the captured hidden-state tensor on the scheduler's GPU and returns per-token logits in place of the hidden states, so the same payload envelope of \S\ref{subsec:hs-payload} carries the result with the shape reduced from $[T, H]$ to $[T]$ and the cross-process hidden-state transfer eliminated entirely.

On the full 144-trajectory \textsc{MiMo-V2-Flash} SWE-QA replay reported in \S\ref{subsec:latency}, the in-engine pruning head adds $8.4\%$ aggregate wall time relative to total generation time. The trade-off is that in-engine colocation modifies the inference engine's request schema and ties the deployment to a specific head architecture, whereas the off-engine path is engine-agnostic and treats the head as an external service.

Two factors will lower the relative overhead reported in \S\ref{subsec:latency}, acting on opposite sides of the ratio. \textbf{Numerator-side}, the pruning cost itself can shrink: the decode path our pruning overhead is measured against has received years of inference-engine optimisation (overlap, paged attention, speculative decoding, prefix caching, custom kernels), whereas our pruning path is currently a hand-tuned baseline of an fp16 base64 envelope and a single-process head; quantising the hidden-state payload to fp8 or INT8 with a custom pack/unpack kernel halves or quarters the bytes on top of the fp16 reduction in Table~\ref{tab:hs-payload}. \textbf{Denominator-side}, modern coding agents increasingly interleave extended reasoning between tool calls, with per-turn thinking budgets now standard practice across frontier coding models; under such heavy-decode workloads per-turn decode dominates wall-clock and the same per-call pruning cost occupies a much smaller fraction of it. These engineering opportunities and heavier-decode workloads may further reduce the relative overhead.

\FloatBarrier
\section{Qualitative cases: per-line head scores on tool outputs}
\label{app:cases}
We sample four cases from the held-out evaluation set (\textsc{Qwen3-Coder-Next} backbone), one for each of the most common non-edit tool families: \texttt{cat}, \texttt{grep}, \texttt{ls}, and test execution.
For each case we show the head's per-line scores alongside the gold annotation.

\subsection{Read case: \texttt{view /testbed/snapshot\_dbg\_cli/\_\_init\_\_.py}}

\begin{figure}[!t]
\centering
\includegraphics[
  max width=0.88\linewidth,
  max totalheight=0.53\textheight
]{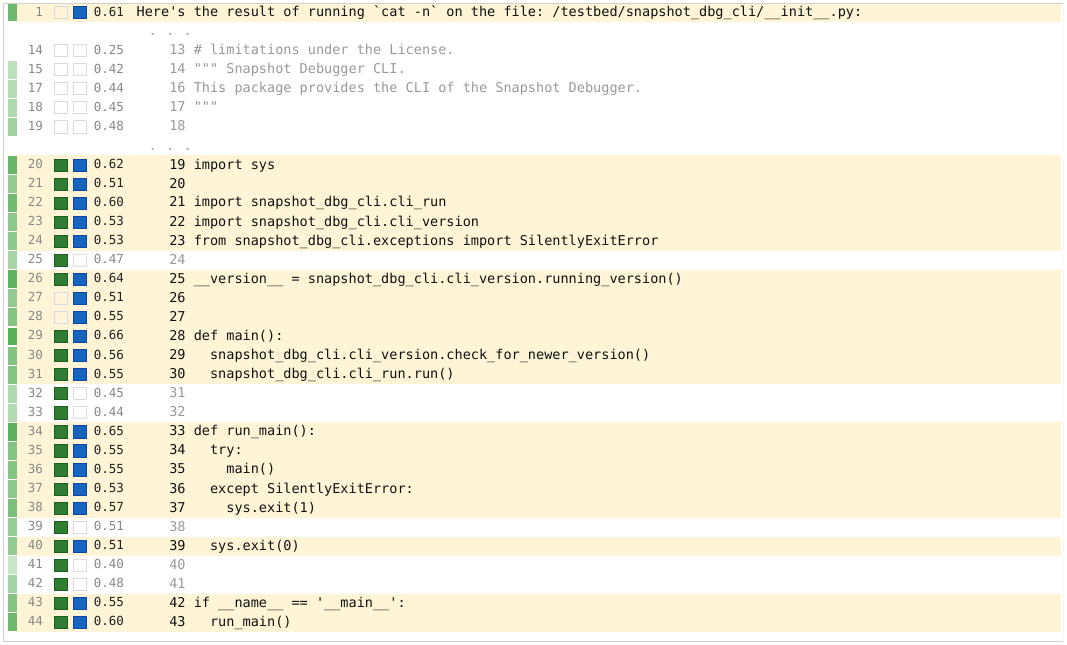}
\caption{Read case: \texttt{view /testbed/snapshot\_dbg\_cli/\_\_init\_\_.py}.
\textbf{Legend (shared by all four cases).}
The left score gutter encodes the head's predicted keep probability $p_i = \sigma(z_i)$, fading from white (low) to \textcolor[HTML]{2e7d32}{\textbf{green}} (high).
\textcolor[HTML]{2e7d32}{\textbf{G}} is filled when the gold annotator marked the line as \emph{keep}; \textcolor[HTML]{1565c0}{\textbf{P}} is filled when the head predicts \emph{keep} at $\tau{=}0.5$.
Predicted-keep rows additionally get a \colorbox[HTML]{fff8c4}{yellow row wash} with crisp black text; pruned rows are dimmed.
\textbf{Case.}
The head assigns the lowest scores in the response (0.25--0.48) to the license/docstring boilerplate---the omitted lines 2--13 and the visible \texttt{\# limitations under the License} plus the package docstring on lines 14--19---while concentrating mass at 0.51--0.66 on the imports, the \texttt{\_\_version\_\_} assignment, and the two function bodies (\texttt{main}, \texttt{run\_main}), exactly the symbols the agent will reference when wiring its patch.
The path-header line~1 is also kept, anchoring the file identity for downstream turns.}
\label{fig:case-read}
\end{figure}

\paragraph{Setup.}
The agent opens a package's \texttt{\_\_init\_\_.py} to learn the CLI entry-point structure before patching it.
The 44-line response is summarised in Figure~\ref{fig:case-read}; gold keeps 23 lines, the head keeps 18 at $\tau{=}0.5$.

\subsection{Search case: \texttt{grep -A 25} for a Flask helper}

\begin{figure}[!t]
\centering
\includegraphics[
  max width=0.88\linewidth,
  max totalheight=0.53\textheight
]{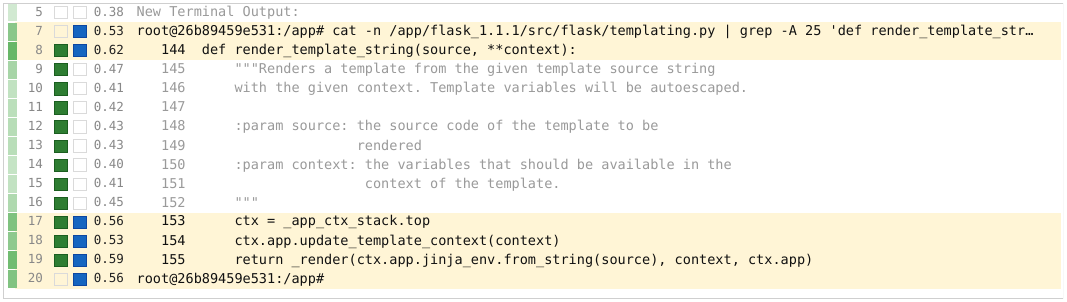}
\caption{Search case: \texttt{cat -n /app/flask\_1.1.1/src/flask/templating.py | grep -A 25 'def render\_template\_string'}.
The head allocates the highest scores to the function's executable body---the \texttt{\_app\_ctx\_stack} lookup, the \texttt{update\_template\_context} call, and the final \texttt{\_render(\dots)} return---which together fully specify how a string template is dispatched.
The intervening docstring lines are conservatively pruned even though the annotator keeps them; the head prioritises the call sequence over the prose as the authoritative spec.}
\label{fig:case-search}
\end{figure}

\paragraph{Setup.}
The agent is investigating Flask's templating layer and pipes \texttt{cat -n templating.py} through \texttt{grep -A 25 'def render\_template\_string'} to view the function and its tail context (Figure~\ref{fig:case-search}).
The 20-line response has gold 12, predicted 6.

\subsection{Listing case: \texttt{ls -la} on legacy admin config dirs}

\begin{figure}[!t]
\centering
\includegraphics[
  max width=0.84\linewidth,
  max totalheight=0.50\textheight
]{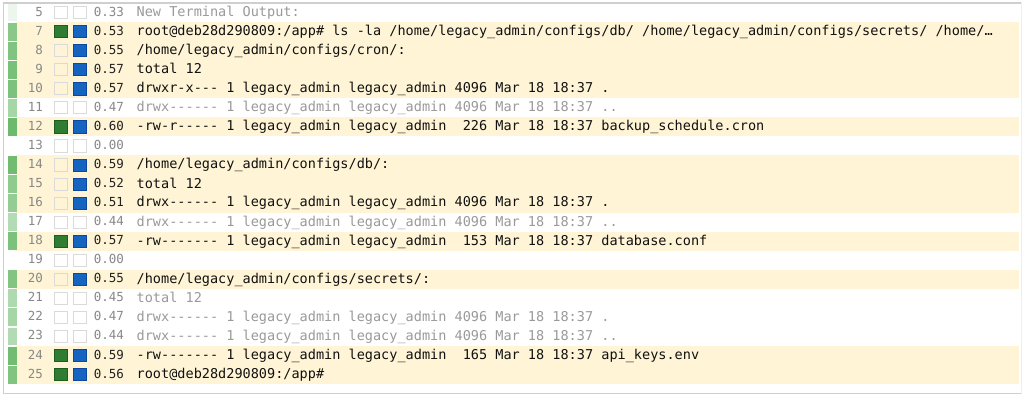}
\caption{Listing case: \texttt{ls -la /home/legacy\_admin/configs/\{cron,db,secrets\}/}.
The head keeps the issued command and every actual file (\texttt{backup\_schedule.cron}, \texttt{database.conf}, \texttt{api\_keys.env}), which are the only artefacts the agent will need to act on.
The remaining eight over-keeps are concentrated on group boundaries rather than scattered noise: the three directory-path headers (\texttt{/cron/:}, \texttt{/db/:}, \texttt{/secrets/:}), two of the three \texttt{total~12} summaries, two leading \texttt{.}~entries, and the closing shell prompt. The head still cleanly prunes the redundant \texttt{..}~parents, blank separators, and the third \texttt{total~12}, so its conservative bias on this listing costs structural boilerplate, not signal.}
\label{fig:case-listing}
\end{figure}

\paragraph{Setup.}
The agent is auditing a legacy admin home and lists three sibling configuration directories (\texttt{cron/}, \texttt{db/}, \texttt{secrets/}) in one call (Figure~\ref{fig:case-listing}).
Of the 25-line response, gold keeps 4 lines and the head keeps 12.

\subsection{Test case: \texttt{python test\_fix.py} traceback}

\begin{figure}[!t]
\centering
\includegraphics[
  max width=0.88\linewidth,
  max totalheight=0.53\textheight
]{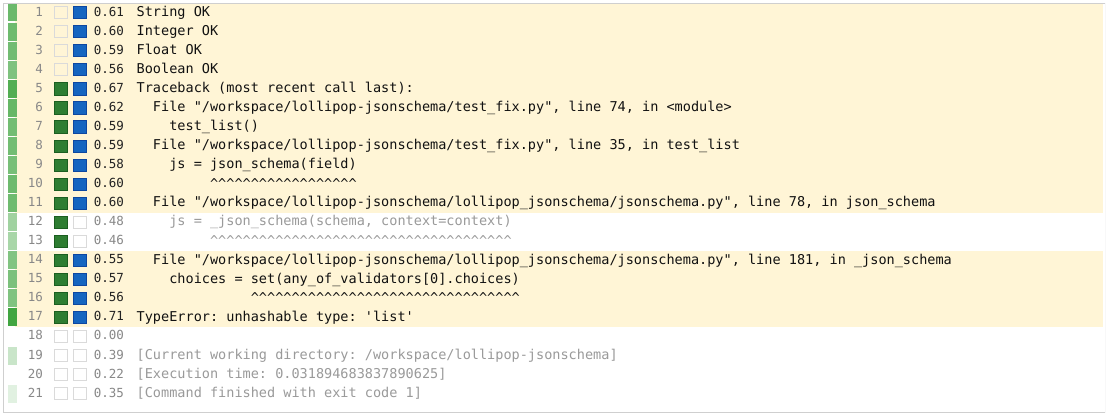}
\caption{Test case: \texttt{python test\_fix.py} (\texttt{lollipop-jsonschema}).
The head locks onto the outer frames of the stack trace: the \texttt{Traceback} header, the \texttt{<module>} and \texttt{test\_list} call sites, the public \texttt{json\_schema} entry, the offending \texttt{set(any\_of\_validators[0].choices)} expression, and the terminal \texttt{TypeError: unhashable type: 'list'} (the highest-scoring line in the response at $0.71$). The intervening inner-dispatch frame (\texttt{js = \_json\_schema(...)} on line~78 of the same file) is conservatively pruned, since it only forwards into the line that already carries the highest keep score.
The four passing lines (\texttt{String OK}, \texttt{Integer OK}, \texttt{Float OK}, \texttt{Boolean OK}) are kept too, providing the contrast that localises the bug to the list path. The trailing harness footer (cwd, exit-code banner) prunes cleanly.}
\label{fig:case-test}
\end{figure}

\paragraph{Setup.}
The agent is debugging a JSON-schema generator (\texttt{lollipop-jsonschema}) and re-runs its repro script after an edit (Figure~\ref{fig:case-test}).
The 21-line tail has gold 13, predicted 15.

\paragraph{Cross-case observation.}
A consistent pattern across the four cases is that per-token sigmoid scores are tightly compressed into a narrow $\sim$0.4--0.7 band rather than saturated near $0$ or $1$.
The head is therefore not learning a single global keep threshold; it produces a fine-grained relative ranking and relies on the length-aware embedding of Eq.~\ref{eq:size-bias} to shift the operating point so the per-line majority vote of Eq.~\ref{eq:majority} lands at the right absolute keep rate for each response length. This matches the design intent of \S\ref{subsec:head}: the head identifies relative importance, and the length-aware embedding absorbs the absolute keep rate that varies with response length.

\FloatBarrier
\section{Qualitative cases: where F1 disagrees with the LLM judge}
\label{app:f1-vs-judge}

The ablation in \S\ref{subsec:ablation} shows that F1 and the LLM judge can diverge sharply on the loss axis: Dice and Tversky reach F1 within 0.04 of our default but their judge scores collapse to $5.30$ and $3.03$, and even BCE outscores our default on F1 in some checkpoints despite being clearly worse to the agent. To see why a high-F1 head can still produce an unusable skeleton, we examine two real tool-response pairs from the held-out evaluation set, each scored against the gold annotation by both metrics in parallel.

\begin{figure}[H]
\centering
\includegraphics[
  max width=0.92\linewidth,
  max totalheight=0.54\textheight
]{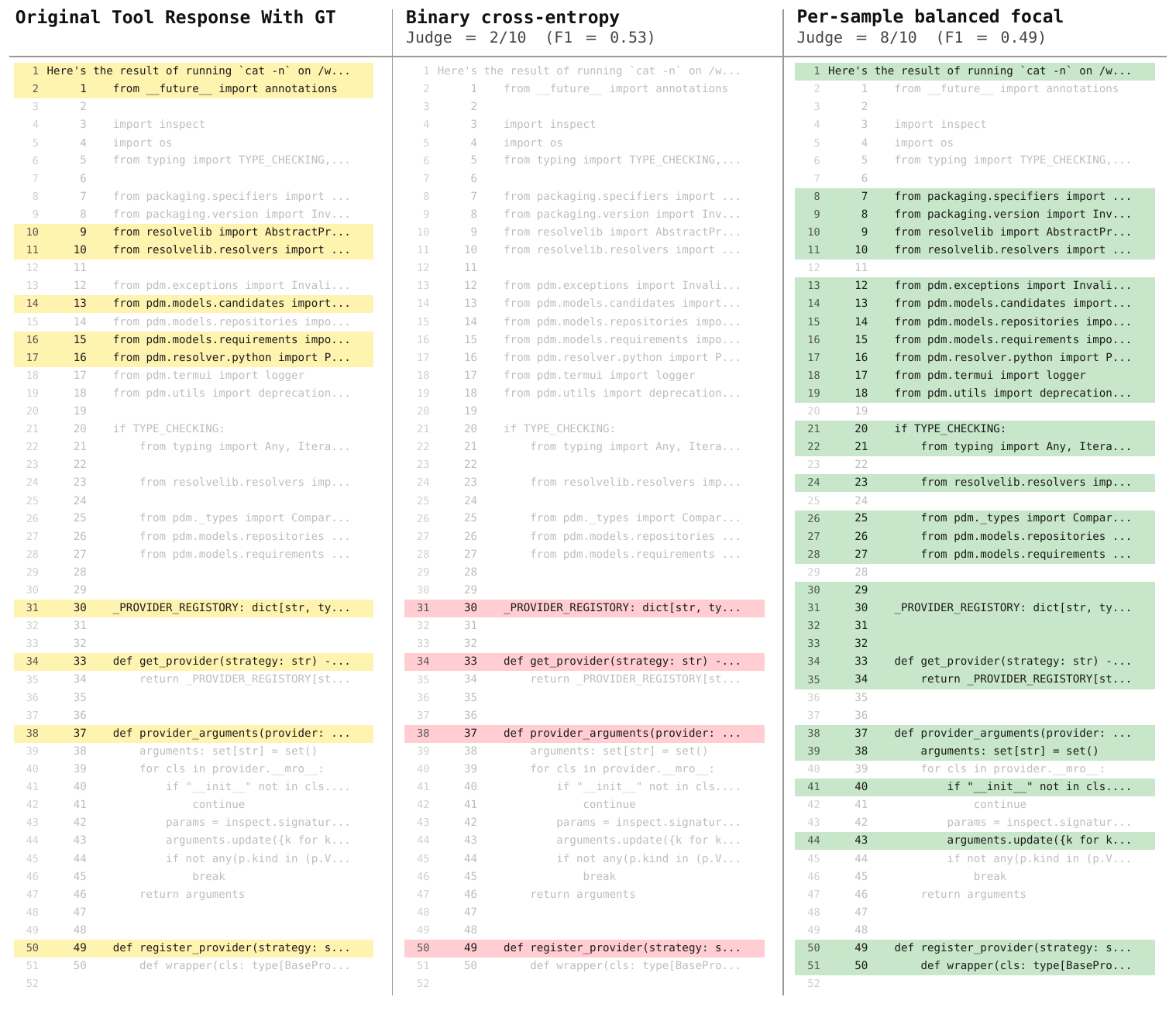}
\caption{\textbf{F1 can rank a useless head above a useful one.} A $52$-line view of \texttt{pdm/resolver/providers.py}; the agent is investigating a recursion bug in \texttt{register\_provider} (L50). Per-sample balanced focal (\textsc{PSBF}, green) keeps $30$ lines covering imports, the \texttt{TYPE\_CHECKING} block, the provider registry, and the bodies of \texttt{get\_provider}, \texttt{provider\_arguments}, and \texttt{register\_provider}; \textsc{BCE} (red) keeps only $4$ isolated lines (L31, L34, L38, L50), namely the variable and function signatures, with no bodies.}
\label{fig:f1-vs-judge-case}
\end{figure}

\begin{figure}[H]
\centering
\includegraphics[
  max width=0.92\linewidth,
  max totalheight=0.54\textheight
]{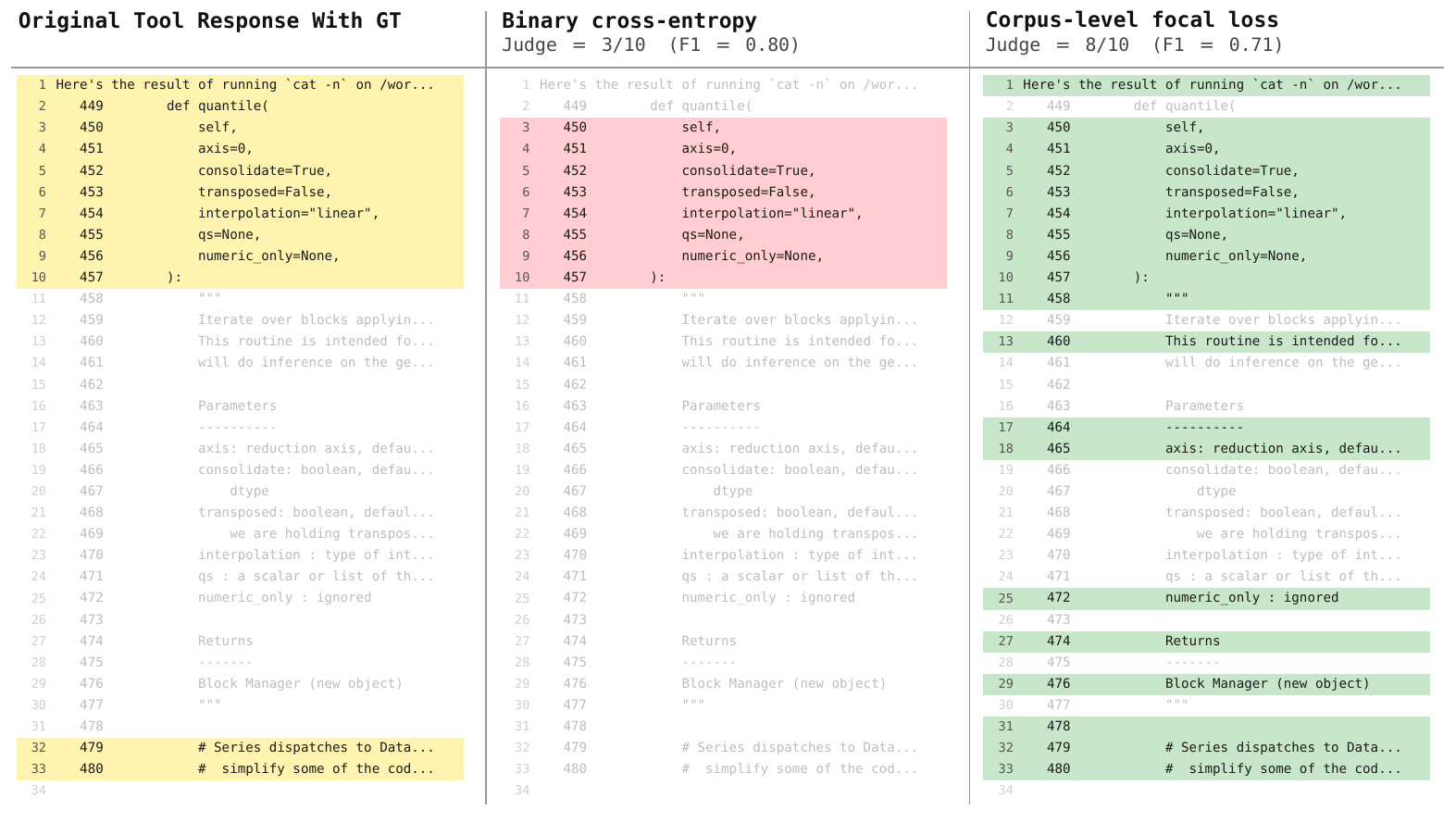}
\caption{\textbf{The same failure mode on a second case.} A $34$-line view of \texttt{pandas/core/internals/managers.py} showing the start of \texttt{quantile()}. \textsc{Focal} (green) keeps the function signature plus the docstring sections explaining reduction semantics and the comment block at the start of the body; \textsc{BCE} (red) keeps only the parameter list (L3--L10) as a contiguous high-precision block with no docstring and no body.}
\label{fig:f1-vs-judge-case-quantile}
\end{figure}

The pattern repeated across both cases is the same. F1 treats the kept set as unweighted membership: a head can attain near-perfect precision by keeping a small high-confidence subset (a function signature or a parameter list, for instance), drop the rest, and be rewarded for the lines it did keep. The judge, in contrast, asks whether the kept skeleton supports the agent's next move; a caller-only or signature-only view that omits bodies, imports, or docstrings receives a low score regardless of how precise the kept lines are. F1 thus rewards picking a tight subset of gold-marked lines, while the judge rewards lines the agent can actually use. The two figures below illustrate this on a recursion-debugging task in \texttt{pdm} (Figure~\ref{fig:f1-vs-judge-case}) and on the entry to \texttt{pandas}'s \texttt{quantile()} (Figure~\ref{fig:f1-vs-judge-case-quantile}).

In Figure~\ref{fig:f1-vs-judge-case}, F1 ranks \textsc{BCE} above \textsc{PSBF} ($0.53$ vs.\ $0.49$) because \textsc{BCE} attains perfect precision on its small kept set, while the judge scores \textsc{PSBF} $8/10$ and \textsc{BCE} $2/10$ ($\Delta=6$). The reason is concrete: a caller-only skeleton with no function bodies and no imports cannot help the agent reason about recursion, so the kept lines are technically correct but operationally useless.

Figure~\ref{fig:f1-vs-judge-case-quantile} shows the same dynamic on a different task. F1 again ranks \textsc{BCE} above \textsc{Focal} ($0.80$ vs.\ $0.71$), yet the judge scores \textsc{Focal} $8/10$ and \textsc{BCE} $3/10$. A parameter list without the docstring's reduction semantics gives an agent only the call interface, not enough to write a correct downstream invocation; the per-line F1 cannot see this gap, but the judge does. Taken together, these cases concretise why we report judge scores alongside F1 in Table~\ref{tab:ablation}: F1 measures label match on the line set, the judge measures usability of the resulting skeleton, and the two can move in opposite directions when a head's precision comes from being narrow rather than from being useful.

\Needspace{0.25\textheight}
\section{Prompts}
\label{app:prompts}


\paragraph{Trajectory labelling prompt.}
Used by \textsc{Claude Sonnet} to annotate per-line keep/prune labels on agent trajectories during dataset construction.

\begin{promptbox}{Trajectory labelling prompt}
SYSTEM_PROMPT = """\
You are an expert at understanding AI coding agents. You specialize in \
compressing tool output so that an agent can still navigate and complete its task. \
Your goal is to produce a filtered view that preserves structural context — \
the agent should be able to orient itself in the codebase after filtering.

You always respond with a single JSON object."""


LABEL_TEMPLATE = """\
An AI coding agent is working on a software engineering task. Below is a window \
from its trajectory: recent conversation context, the current tool call + response, \
and what the agent does next.

Your job: decide which lines to KEEP so the agent can still work effectively. \
The lines you remove will be replaced by "(filtered N lines)" markers in the output \
the agent sees. Think of yourself as producing a **skeleton view** — the agent \
should still be able to locate code, understand structure, and find what it needs.

## Previous conversation (context):

{history}

## Current tool call:

Tool: {tool_name}
Arguments: {tool_args}

## Tool response ({n_lines} lines):

Each line is prefixed with "L<number>| " — these are the OUTER line numbers you must use.

{numbered_code}

## What the agent does next:

{next_turn}

---

**CRITICAL**: The tool response may contain internal line numbers (e.g., from `cat -n` \
or editor output showing file line numbers like `    60\tdef foo():`). \
**IGNORE those internal numbers.** Only cite the OUTER line numbers (the "L1|", "L2|", \
"L3|"... prefixes added by this system). For example, if L3 shows `    60\tdef foo():`, \
cite [3], NOT [60].

Decide which lines to KEEP. The kept lines should form a **readable skeleton** of \
the original output. After filtering, the agent will see something like:

```
def authenticate(request):
    token = request.headers.get("Authorization")
(filtered 12 lines)
    if not user.is_active:
        raise PermissionError("Account disabled")
(filtered 8 lines)
def logout(request):
(filtered 5 lines)
```

### What to KEEP:

1. **Lines the agent directly uses next** — code it edits, references, or reasons about
2. **Structural boundaries** — function/class/method signatures, decorator lines, \
   closing braces/brackets. When keeping any line inside a block, ALWAYS keep the block's \
   opening signature (def/class/if/for/try/with). This is the most important rule — \
   the agent needs these landmarks to navigate
3. **Key definitions** — imports, variable assignments, type declarations that the \
   agent's focus depends on
4. **Error-relevant lines** — stack traces, error messages, assertion failures, \
   test names with PASS/FAIL status
5. **Section headers** — file paths, separators, command output markers that help \
   the agent orient in long output

### What to REMOVE:

- Blank lines, pure comment blocks, license headers
- Function bodies that are unrelated to the agent's current focus
- Repetitive output (e.g., long lists where a few examples suffice)
- Verbose boilerplate (import blocks where only 1-2 are relevant)

### Confidence:

- **confident** — You can clearly identify which lines matter from context + next action.
- **skeleton** — You're unsure which specific lines the agent will focus on (e.g., the \
  next action targets a different file, the context is ambiguous, or multiple very \
  different subsets seem equally valid). Keep only structural skeleton: \
  function/class signatures, import lines, section boundaries. Strip body details.

Think step-by-step before producing the JSON:

1. What is the agent trying to accomplish?
2. What does it do next with this output?
3. Which lines does it directly need?
4. Which structural boundaries (function/class signatures) should remain as landmarks?
5. Imagine the filtered output with "(filtered N lines)" gaps — can the agent still work?
6. Am I confident about which specific content lines matter, or should I fall back to skeleton?

Respond with a single JSON object (no markdown fences):
{{
  "reasoning": "<1-2 sentences: what the agent is doing and why these lines matter>",
  "confidence": "confident" or "skeleton",
  "kept_lines": [1, 3, "5-7", "20-30"]
}}

kept_lines supports single numbers and "start-end" range strings."""
\end{promptbox}

\Needspace{0.16\textheight}\paragraph{Architecture-ablation judge prompt.}
Used by \textsc{GPT-5.4-mini} to score pruner output quality on a $1$--$10$ rubric.

\begin{promptbox}{Architecture-ablation judge prompt}
JUDGE_SYSTEM = """\
You are an expert at evaluating AI coding agent tool response pruning.

A pruner filters tool responses to keep only a **readable skeleton** — the \
minimal set of lines that lets the agent proceed identically. An ideal pruned \
response looks like:

```
def authenticate(request):
    token = request.headers.get("Authorization")
(filtered 12 lines)
    if not user.is_active:
        raise PermissionError("Account disabled")
(filtered 8 lines)
def logout(request):
(filtered 5 lines)
```

### Lines that SHOULD be kept:

1. Lines the agent directly uses next — code it edits, references, or reasons about
2. Structural boundaries — function/class/method signatures, closing braces/brackets
3. Key definitions — imports, variable assignments the agent's focus depends on
4. Error-relevant lines — stack traces, error messages, test PASS/FAIL status
5. Section headers — file paths, separators, command output markers

### Lines that SHOULD be removed:

- Blank lines, pure comment blocks, license headers
- Function bodies unrelated to the agent's current focus
- Repetitive output (long lists where a few examples suffice)
- Verbose boilerplate (import blocks where only 1-2 are relevant)

The "Agent's next action" shows what the agent did after receiving the \
**original** (unpruned) response. Use it to understand what information \
the agent actually needed from the tool response.

Score on TWO dimensions, then combine:

- **Recall**: Does the pruned version retain all lines the agent needs?
- **Precision**: Does it remove lines the agent does NOT need?

Score 1-10 and give a brief reason. Output a JSON object:
{"score": <int 1-10>, "reason": "<1-2 sentences>"}

Scoring guide:

- **9-10**: Near-ideal skeleton. All critical lines kept, noise removed. \
  Agent could proceed identically with a compact response.
- **7-8**: Good skeleton but minor issues — a few useful lines missing, \
  or some unnecessary lines kept.
- **5-6**: Mediocre. Either missing important lines OR keeping too much \
  (>80% kept with obvious removable noise).
- **3-4**: Poor. Significant information lost, OR essentially no pruning \
  done (>90% kept on a response with clear boilerplate).
- **1-2**: Useless. Critical information destroyed, or all lines pruned."""
  \end{promptbox}

\paragraph{SWE-QA family answer-quality judge prompt.}
Used by \textsc{GPT-5.4-mini} to score the agent's final answer against a reference answer on a $1$--$10$ rubric across five dimensions (correctness, completeness, relevance, clarity, reasoning).
\begin{promptbox}{SWE-QA family answer-quality judge prompt}
SCORING_PROMPT = """\
You are an expert evaluator. Compare a candidate answer against a reference answer for a code repository question.

Question: {question}

Reference Answer:
{reference}

Candidate Answer:
{candidate}

Score the candidate on these 5 dimensions (1-10 each):

1. correctness: Are the core facts and details accurate?
2. completeness: Does it cover all key points from the reference?
3. relevance: Is it focused on the question without irrelevant information?
4. clarity: Is the language clear and precise?
5. reasoning: Is the reasoning logical and well-structured?

Respond with ONLY a JSON object (no explanation):
{{"correctness": N, "completeness": N, "relevance": N, "clarity": N, "reasoning": N}}"""

\end{promptbox}

\end{document}